\documentclass[twoside]{article}

\usepackage{amsmath}
\usepackage{amssymb}
\usepackage{mathtools}
\usepackage{amsthm}

\usepackage{microtype}
\usepackage{graphicx}
\usepackage{wrapfig}
\usepackage{subfig}

\usepackage{float}

\usepackage{booktabs} 
\usepackage{xspace}
\usepackage{hyperref}
\usepackage[inline]{enumitem}
\usepackage{algorithm}
\usepackage{algorithmic}
\usepackage{varwidth}
\usepackage[utf8]{inputenc} 
\usepackage[T1]{fontenc}    
\usepackage{hyperref}       
\usepackage{url}            
\usepackage{booktabs}       
\usepackage{amsfonts}       
\usepackage{nicefrac}       
\usepackage{microtype}      
\usepackage{xcolor}         
\usepackage{ragged2e}
\newcommand{\eqnote}[2][0.32\linewidth]{\parbox{#1}{\footnotesize\upshape\RaggedRight#2}}

\newtheorem{assumption}{Assumption}
\newtheorem{theorem}{Theorem}

\usepackage[accepted]{aistats2025}
\usepackage[round]{natbib}

 \DeclareMathOperator*{\E}{\mathbb{E}}

\newcommand{\pie}{\pi_e}

\DeclarePairedDelimiter\abs{\lvert}{\rvert}%
\DeclarePairedDelimiter\norm{\lVert}{\rVert}%

\newcommand{\renyi}{R\'enyi divergence\xspace}

\newcommand{\task}{\mathcal{T}_{\text{ope}}}

\makeatletter
\let\oldabs\abs
\def\abs{\@ifstar{\oldabs}{\oldabs*}}
\let\oldnorm\norm
\def\norm{\@ifstar{\oldnorm}{\oldnorm*}}
\makeatother

\newcommand{\bea}{\begin{eqnarray}} 
\newcommand{\eea}{\end{eqnarray}}

\newcommand{\bbAutoOPE}{\mbox{AutoOPE}\xspace}
\newcommand{\bbextAutoOPE}{\mbox{Automated} \mbox{Off-Policy} \mbox{Estimator} \mbox{Selection}\xspace}

\newcommand{\fakecite}[1]{\ignorespaces} 
\newcommand{\ie}{i.e.,\xspace}
\newcommand{\eg}{e.g.,\xspace}
\newcommand{\pasif}{\mbox{PAS-IF}\xspace}

\newcommand{\nico}[1]{\ignorespaces}
\newcommand{\mf}[1]{\ignorespaces}
\newcommand{\michael}[1]{\ignorespaces}

\begin{document}

%

%

\twocolumn[

\aistatstitle{Automated Off-Policy Estimator Selection via Supervised Learning}

\aistatsauthor{ Nicolò Felicioni \And Michael Benigni \And  Maurizio Ferrari Dacrema}

\aistatsaddress{ Politecnico di Milano \\ {\tt nicolo.felicioni@polimi.it} \And  Politecnico di Milano \\ {\tt michael.benigni@polimi.it} \And Politecnico di Milano \\ {\tt maurizio.ferrari@polimi.it} } ]

\begin{abstract}
The Off-Policy Evaluation (OPE) problem consists of evaluating the performance of counterfactual policies with data collected by another one. 
To solve the OPE problem, we resort to estimators, which aim to estimate in the most accurate way possible the performance that the counterfactual policies would have had if they were deployed in place of the logging policy. In the literature, several estimators have been developed, all with different characteristics and theoretical guarantees. Therefore, there is no dominant estimator and each estimator may be the best for different OPE problems, depending on the characteristics of the dataset at hand.
Although the selection of the estimator is a crucial choice for an accurate OPE, this problem has been widely overlooked in the literature. We propose an automated data-driven OPE estimator selection method based on supervised learning. In particular, the core idea we propose in this paper is to create several synthetic OPE tasks and use a machine learning model trained to predict the best estimator for those synthetic tasks. We empirically show how our method is able to perform a better estimator selection compared to a baseline method on several real-world datasets, with a computational cost significantly lower than the one of the baseline.
\end{abstract}

\section{INTRODUCTION}
\label{sec:introduction}
In numerous real-world scenarios, decision-making systems are required to interact with dynamic environments, making choices aimed at optimizing specific objectives. These systems are prevalent in domains such as recommender systems \citep{gilotte, saito-johachins-21, DBLP:conf/nips/FelicioniDRC22}, information retrieval \citep{DBLP:conf/www/LiCKG15}, ad-placement systems \citep{DBLP:journals/jmlr/BottouPCCCPRSS13}, or medical treatments \citep{gan2023contextual}. Such decision-making problems can be viewed through the lens of the \textit{contextual bandit} paradigm \citep{cmab1, cmab2}. For instance, let us consider the recommendation system domain: the recommender (agent) repeatedly observes user profiles (contexts), and aims to recommend items (actions) to users based on the given context to maximize some notion of reward.
A critical challenge arises in evaluating the performance of new policies without deploying them in the real world. This is especially pertinent in scenarios where deploying a suboptimal policy can have significant negative consequences. \textit{Off-Policy Evaluation} (OPE) addresses this challenge by evaluating the performance of counterfactual policies using data collected from a different, previously deployed policy, often termed the \textit{logging} policy. The ability to make informed decisions about policy modifications without necessitating real-world deployment is not only cost-effective but also reduces the potential risks associated with untested policies.

However, the effectiveness of OPE largely hinges on the accuracy of the \textit{estimator} used to predict the performance of the \textit{evaluation} policies. 
There is a wide variety of estimators that have been proposed in the literature, each with distinct characteristics and mathematical properties \citep{vlassis2019design}. As an example, a well-known estimator, called \textit{Inverse Propensity Scoring} (IPS) \citep{ips1, ips2}, provides an unbiased estimation, at the cost of high variance.
Other estimators also exist, each with unique trade-offs between bias and variance. 
Given the heterogeneity of real-world datasets and the myriad of scenarios in which OPE is applied, no single estimator is universally optimal. 
An estimator may be the best under certain circumstances but perform poorly under others. This variety of estimators, while offering flexibility, also introduces the challenge of \textit{Estimator Selection}, which is a crucial problem in OPE and has been mostly overlooked in the existing literature.
Basically, the Estimator Selection problem consists of the following: we are given an Off-Policy Evaluation task and we have to select the best estimator for the task we have at hand. 

In this paper, we address the Estimator Selection problem using an automated data-driven approach. 
Our main idea consists of framing this problem as a supervised learning problem. In particular, we notice that there are several results suggesting that an estimator's error is related to the OPE task characteristics. Hence, we generate several different synthetic OPE tasks by varying the characteristics of each one, we apply various types of estimators and their estimates are compared against the ground-truth value of each evaluation policy, which is known due to the synthetic nature of the data. 
These synthetic OPE tasks are used to train a supervised model designed to predict estimator's Mean Squared Error for the given OPE task. 
The supervised model will understand the underlying patterns shared among the various OPE tasks and their relation with the estimator error, and then it can be used to select the estimator with the lowest predicted error, effectively addressing the OPE Estimator Selection challenge. We call our Estimator Selection method \textit{\bbextAutoOPE} (\bbAutoOPE). 
The problem of Estimator Selection has been widely overlooked in the literature, and, to the best of our knowledge, only a very recent paper \citep{saito23} proposed a method for Estimator Selection, named Policy-Adaptive Estimator Selection via Importance Fitting (\pasif), which is substantially different from our proposal. 


In this paper, we propose a novel data-driven approach called \bbextAutoOPE (\bbAutoOPE) to address the Estimator Selection problem in Off-Policy Evaluation. Our key contributions are:
\begin{itemize}[noitemsep]
    \item We train a machine learning model to predict the performance of estimators on a large number of synthetic OPE tasks with varying characteristics. We share the code for our experiments for researchers and practitioners in the field\footnote{We anonymously share the code to reproduce our experiments: \url{https://anonymous.4open.science/r/auto-ope-28F3/}}.
    \item To the best of our knowledge, this is 
    one of the few techniques proposed for Estimator Selection, despite the importance of the problem. 
    \item We conduct extensive experiments on several datasets that empirically demonstrate the effectiveness of our proposed \bbAutoOPE method in selecting high-performing estimators tailored to the given OPE task, performing better than the baseline with a much reduced computational cost. 
\end{itemize}


\section{BACKGROUND}
\label{sec:background}

\subsection{Contextual Bandit Problem}
\label{subsec:contextual-bandits}
The Contextual Bandit (CB) problem is an instance of sequential decision-making problems. At a given time instant $t$, an agent observes a \textit{context} $x_t \sim p(\cdot)$ from a space $\mathcal{X}$, and chooses an \textit{action} $a_t$ from a set $\mathcal{A}$, according to a \textit{policy} $\pi_b$, that is a statistical distribution of the actions conditioned on the context ($a_t \sim \pi_b(\cdot|x_t)$), called \textit{logging policy}. 
After choosing an action $a_t$ the agent receives a \textit{reward} $r_t$ from the environment, sampled from $r_t \sim p(\cdot|x_t,a_t)$. This means that the reward observed is conditioned on both context and action.
A common quantity used to compare the quality of different policies is the \textit{policy value}, that is defined as the expected value of the reward collected under a certain policy $\pi$: $ V(\pi) \coloneqq \mathbb{E}_{p(r_t|x_t, a_t)\pi(a_t|x_t)p(x_t)}\left[r_t\right] $.

\subsection{Off-Policy Evaluation}
\label{subsec:ope}
The Policy Evaluation problem refers to the task of estimating the policy value of a given policy $\pi_e$, called \textit{evaluation policy}, and understanding how it would perform in a given environment. 
In this paper we focus on \textit{Off-Policy Evaluation} (OPE). 
In OPE, we would like to estimate the policy value of the evaluation policy $\pi_e$ using data collected under a different logging policy $\pi_b$. This logging data is defined as $\mathcal{D}_b = \left\{\left(x_t, a_t, r_t, \left\{\pi_b(a|x_t)\right\}_{a \in \mathcal{A}}\right)\right\}_t^{n_b}$, with $n_b$ being the number of tuples in $\mathcal{D}_b$, also called number of rounds. We define an OPE \textit{task} $\task$ as the set composed of the logging dataset and the evaluation policy:
\begin{align}
    \task \coloneqq \left\{\mathcal{D}_b, \left\{\pi_e(a|x_t)\right\}_{a \in \mathcal{A}}\right\}
    \label{eq::D-ope}
\end{align}

The problem of Off-Policy Evaluation can be tackled with OPE \emph{estimators}, 
which can be defined as a statistical function that uses logging data $\mathcal{D}_b$ collected under the logging policy $\pi_b$ 
as well as the evaluation policy $\pi_e$ computed for each context $x_t$, 
to estimate the ground-truth evaluation policy value $V(\pi_e)$.
We denote an OPE \textit{estimator} as $\hat{V}(\pi_e)$.


A common metric used to assess the performance of an estimator is the Mean Squared Error (MSE), defined as $\text{MSE}(\hat{V}(\pi_e)) = \mathbb{E} \left[(\hat{V}(\pi_e) - V(\pi_e))^2 \right]$.

\subsection{Estimator Selection for OPE}
\label{subsec:op-estimator-selection}

The effectiveness of OPE crucially depends on how accurate is the estimator in estimating the policy value of the evaluation policy, usually in terms of how low is its $\text{MSE}(\hat{V}(\pi_e))$.
For this reason, and since many OPE estimators have been proposed in the literature, selecting the best estimator for a given OPE task, \ie \textit{Estimator Selection}, is particularly important. Given a finite set $\mathcal{V}$ of different OPE estimators, the Estimator Selection problem consists in finding the best estimator among them. 

Since in OPE we do not have any data collected under the evaluation policy, we cannot compute the ground-truth evaluation policy value $V(\pi_e)$ and therefore the ground-truth MSE of the estimator. A common approach is to approximate it with an estimate, \ie $\widehat{\text{MSE}}(\hat{V}(\pi_e))$, and to select the estimator that has the lowest approximate MSE.

\subsubsection{Policy-Adaptive Estimator Selection via Importance Fitting (\pasif)}
\label{subsec:pasif}
To the best of our knowledge, only \textit{Policy-Adaptive Estimator Selection via Importance Fitting} (\pasif) \citep{saito23} has tackled the problem of Estimator Selection in the literature. 
To accurately estimate the MSE of an OPE estimator, the key idea of \pasif is to determine a sampling rule $\rho_\theta: \mathcal{X} \times \mathcal{A} \rightarrow (0, 1)$, that governs a subsampling procedure to divide the logging data $\mathcal{D}_b$ into two \textit{pseudo datasets}: a pseudo logging dataset $\tilde{\mathcal{D}}_b$ and a pseudo evaluation dataset $\tilde{\mathcal{D}}_e$. 
$\rho_\theta(x_t,a_t)$ represents the probability of assigning the logging sample $(x_t, a_t)$ to $\tilde{\mathcal{D}}_e$. In this context, the two pseudo datasets can be viewed as if they were collected by two \textit{pseudo policies}, $\tilde{\pi}_b$ and $\tilde{\pi}_e$, induced by $\rho_\theta$.
The $\rho_\theta(x_t, a_t)$ is the output of a fully connected neural network, optimized by minimizing the average squared error between the importance ratio $w(x_t, a_t) = \pi_e(a_t|x_t) / \pi_b(a_t|x_t)$ induced by true policies, and $\tilde{w}(x_t, a_t) = \tilde{\pi}_e(a_t|x_t) / \tilde{\pi}_b(a_t|x_t)$ induced by the pseudo policies: this learning phase is called \textit{importance fitting}. 
The pseudo evaluation dataset $\tilde{\mathcal{D}}_e$ is deemed to be collected by the evaluation policy, and in this way it can be used to estimate the evaluation policy value by averaging the rewards in it (we denote this estimate as $\hat{V}_{on}(\pi_e; \mathcal{\tilde{D}}_{e})$). The pseudo logging dataset $\tilde{\mathcal{D}}_b$ is used to compute the candidate OPE estimator prediction ($\hat{V}(\pi_e; \mathcal{\tilde{D}}_b)$). After that, each estimator MSE can be approximated in this way:
\begin{align*}
    \widehat{\text{MSE}}_{\pasif} (\hat{V}(\pi_e))  \coloneqq (\hat{V}_{on}(\pi_e; \mathcal{\tilde{D}}_{e}) - \hat{V}(\pi_e; \mathcal{\tilde{D}}_b))^2 \ .
\end{align*}


One limitation of \pasif is that $\rho_\theta(x_t,a_t)$ is different for every OPE task, hence requiring to re-train a separate \pasif model for each OPE task. 
A second limitation is that the ability of \pasif to build two effective pseudo datasets is constrained by the available training data which may be problematic in scenarios where the data is limited.



\section{AUTOMATED OFF-POLICY ESTIMATOR SELECTION}
\label{sec:bb-method}

In this section, we present a novel approach for solving the Estimator Selection problem, called \textit{\bbextAutoOPE} (\bbAutoOPE).

\subsection{Theoretical Motivation}
\bbAutoOPE is motivated by many findings that show how the performance of an OPE estimator is highly dependent on the characteristics of the OPE task at hand \cite{voloshin, robustness-off-policy}. To make one example, 
\citet{metelli-pois} show that the IPS error is strictly related to the (exponentiated) \textit{\renyi} (with $\alpha=2$) between the logging and evaluation policies\footnote{The \renyi is a family of divergences between two probability distributions with a parameter $\alpha$. For $\alpha=2$, it is defined as
$ d_r(\pie || \pi_b ) = \log \E_x \E_{\ a \sim \pi_b} \left[ ( \pie(a|x) / \pi_b(a|x))^2 \right] $.}
and provides a finite sample bound on the absolute estimation error of IPS under standard assumptions:
$\abs{\Hat{V}_{\text{IPS}}(\pie) - V(\pie) } \leq \mathcal{O}\left( \sqrt{\frac{\exp (d_r(\pie || \pi_b ))}{n_b}} \right). $
This theoretical finding is just one example among many others that shows how the performance of an OPE estimator is related to the characteristics of the OPE task. 
In other words, we make the following assumption:
\begin{assumption}
\label{ass:mse-func}
Consider a fixed set of estimators $\mathcal{V}$. There exists a function $f^* \in \mathcal{F}$ that expresses the relationship between the real error of any estimator on a given OPE task $MSE(\hat{V}(\pi_e); \task)$ and the features $g(\hat{V}, \task)$ of the estimator and the task:
\[
\text{MSE}(\hat{V}(\pi_e); \task) = f^*(g(\hat{V}, \task)), \quad \forall \hat{V} \in \mathcal{V} .\]

\end{assumption}

Let us assume that we have a variety of tasks at our disposal $\mathcal{M} \coloneqq \{\task^{(1)}, \task^{(2)}, \dots\}$, a fixed set of estimators $\mathcal{V}$, and the feature extraction function $g$. 
With these, we can frame a supervised learning problem and apply empirical risk minimization to find the function that best predicts the MSE for all tasks and estimators:
\begin{equation*}
    \begin{split}
       \hat{f} = & 
\arg \min_{f \in \mathcal{F}}  \frac{1}{
N
} \sum_{\substack{\Hat{V} \\ \task}}  \mathcal{L} \Bigl( \text{MSE}(\hat{V}(\pi_e); \task) , 
f(g(\hat{V}, \task)) \Bigr), 
    \end{split}
\end{equation*} 
where $\mathcal{L}$ is a suitable loss, $\mathcal{F}$ is the hypothesis class, and $N = |\mathcal{V}||\mathcal{M}| $. Now, we can prove that the empirical risk minimizer $\hat{f}$ is accurate with high probability. To provide an intuition, we consider a finite hypothesis class, but one can extend this finding to infinite hypothesis classes via standard learning theory tools (e.g., VC-dimension, Rademacher complexity) \citep{shalev2014understanding}. Also, we consider a subgaussian loss, which is less restrictive than only considering bounded loss functions \citep{pac-bayesian}.


\begin{assumption}[Sub-Gaussian Loss]
\label{ass:subgaussian-loss}
The loss function $\mathcal{L}$ is subgaussian with parameter $\sigma^2$. Specifically, for all $y, \hat{y}$, the random variable $\mathcal{L}(y, \hat{y}) - \mathbb{E}[\mathcal{L}(y, \hat{y})]$ satisfies:
\[
\mathbb{E}\left[ e^{\lambda (\mathcal{L}(y, \hat{y}) - \mathbb{E}[\mathcal{L}(y, \hat{y})])} \right] \leq e^{\frac{\lambda^2 \sigma^2}{2}} \quad \forall \lambda \in \mathbb{R}.
\]
\end{assumption}

\begin{assumption}[Finite hypothesis class]
\label{ass:finite-hyp}
The hypothesis class $\mathcal{F}$ is finite: $|\mathcal{F}| < \infty$.
\end{assumption}

\begin{theorem}
\label{thm:erm-bound}
Under Assumptions \ref{ass:mse-func}, \ref{ass:subgaussian-loss}, and \ref{ass:finite-hyp}, suppose we have $N = |\mathcal{V}| |\mathcal{M}|$ independent samples $\{(x_i, y_i)\}_{i=1}^N$, where each $x_i = g(\hat{V}_i, \mathcal{T}_i)$ and $y_i = MSE(\hat{V}_i(\pi_e); \mathcal{T}_i)$, drawn uniformly over the estimators $\hat{V} \in \mathcal{V}$ and tasks $\mathcal{T} \in \mathcal{M}$. Let $\hat{f}$ be the empirical risk minimizer over $\mathcal{F}$:
\[
\hat{f} = \arg \min_{f \in \mathcal{F}} \frac{1}{N} \sum_{i=1}^N \mathcal{L}\left( y_i, f(x_i) \right).
\]
Then, for any $\delta > 0$, with probability at least $1 - \delta$, the expected loss $L(\hat{f})$ satisfies:
\[
L(\hat{f}) \leq \sqrt{ \frac{8 \sigma^2 \log \left( \dfrac{2 |\mathcal{F}|}{\delta} \right)}{N} },
\]
where $\sigma^2$ is the subgaussian parameter from Assumption \ref{ass:subgaussian-loss}, and the expected loss $L(f)$ is defined as
\[
L(f) = \mathbb{E}_{(x, y)} \left[ \mathcal{L}\left( y, f(x) \right) \right].
\]
\end{theorem}
\begin{proof}
The proof derives from mapping the estimator selection problem to a supervised learning one, and then applying standard learning theory tools, such as Hoeffding's inequality and the union bound. See Appendix \ref{app:proofs} for all the deferred proofs.
\end{proof}

Theorem \ref{thm:erm-bound} shows that, with high probability, the empirical risk minimizer $\hat{f}$ has a low error, and the error term that decreases with $N$. Since the number of estimator is fixed, the error term decreases the number of tasks $|\mathcal{M}|$ contained in the dataset. 

This theoretical finding is intuitive, and it shows that $\hat{f}$ is trained to understand the patterns of the given tasks and it learns more if provided with more tasks. Once we have found $\hat{f}$, we can apply it in a zero-shot way whenever we need to perform estimator selection with a new task $\task^{(\text{new})} \notin \mathcal{M}$. With $\hat{f}$ we can now estimate the MSE for a given task in this way: 
\[\widehat{MSE}_{\bbAutoOPE}(\hat{V}, \task^{(\text{new})}) \coloneqq \hat{f}(g(\hat{V}, \task^{(\text{new})})).\]
Hence, we just need to select the estimator that minimizes the MSE estimated by \bbAutoOPE: 
\[\hat{V}^* = \arg \min_{\hat{V} \in \mathcal{V}} \widehat{MSE}_{\bbAutoOPE}(\hat{V}, \task^{(\text{new})}) .\]

\subsection{Practical Implementation}

In the previous section, we provided a theoretical intuition for our approach. In this section, we describe how to implement the proposed solution in practice. In particular, this approach needs several OPE tasks, a feature extraction function, and the ground-truth MSE. For this reason, we rely on \textit{synthetic data}. Specifically:

    \paragraph{Dataset} We generate a dataset $\mathcal{M} \coloneqq \{\task^{(1)}, \task^{(2)}, \dots\}$ composed of several different synthetic OPE tasks. In this way, we have access to the ground-truth MSE. Also, since the objective is to obtain a model that is able to generalize in a zero-shot way, we want to train our model with a large number of OPE tasks. This would be unfeasible using real-world OPE data, which is scarce. We indeed generate 250,000 synthetic OPE tasks. Furthermore, to favor generalization of our model, we create OPE tasks with different distributions. For more details on the synthetic data generation, we refer to Appendix \ref{app:bb-dataset}.
    \paragraph{Features} Regarding the features $g$, while theoretical results have proven a relation, for instance, with the exponentiated \renyi, we want to take into account a wide selection of possible features. Hence, we designed a total of 43 features. 
    In particular, we categorize the features into three groups:
    \begin{enumerate*}[label=(\roman*)]
        \item Policies-independent features, that describe structural properties of the CB scenario and are unrelated from the specific $\pi_b,\ \pi_e$, \eg the number of rounds, the reward variance, the number of actions.
        \item Policies-dependent features, such as the dissimilarity between the two policies $\pi_b,\ \pi_e$, measured with different statistical distances and divergences.
        \item OPE estimator features, that refer to specific properties of the OPE estimator used in that task, \eg if the estimator is self-normalized, or uses importance sampling. 
    \end{enumerate*}
    See Appendix \ref{app:feat-eng} for their full description. 
    \paragraph{Estimators} The set $\mathcal{V}$ of candidate estimators include several model-free, model-based and hybrid estimators, as done in \cite{saito23}. All the estimators used are described in Appendix \ref{app:ope-est}. 
    \paragraph{Model} As a supervised model $\hat{f}$, we use a \textit{Random Forest} regressor \citep{random-forest} because of its high-quality predictions on tabular data and its speed, both in training and inference. For more details on training and optimization, we refer to Appendix \ref{app:bb-model}.

\begin{algorithm}[t]
\small
\caption{\bbAutoOPE training}
\begin{algorithmic}[1]
\REQUIRE Feature generator $g$, estimator set $\mathcal{V}$
\ENSURE MSE estimator $\widehat{MSE}_{\bbAutoOPE}$
\STATE Generate dataset  $\mathcal{M} \coloneqq \{\task^{(1)}, \task^{(2)}, \dots\}$ (Appendix  \ref{app:bb-dataset})
\STATE $\mathcal{L} \gets$ Random Forest regression loss 
\STATE $\hat{f} \gets \arg \min_{f}$ \\ $ \sum_{\task \in \mathcal{M}, \Hat{V} \in \mathcal{V}} \mathcal{L}\left( MSE(\hat{V}(\pi_e); \task) , f(g(\hat{V}, \task)) \right)$ (Train the random forest)
\STATE For any $\hat{V}, \task$, define \\ $\widehat{MSE}_{\bbAutoOPE}\left(\hat{V}, \task\right) \coloneqq \hat{f}(g(\hat{V}, \task))$
\RETURN $\widehat{MSE}_{\bbAutoOPE}$
\end{algorithmic}
\label{alg:auto-ope-training}
\end{algorithm}

\begin{algorithm}[t]
\small
\caption{\bbAutoOPE estimator selection}
\begin{algorithmic}[1]
\REQUIRE New OPE task $\task^{(\text{new})}$, estimator set $\mathcal{V}$, MSE estimator $\widehat{MSE}_{\bbAutoOPE}$
\ENSURE Selected estimator $\hat{V}^*$
\STATE $\hat{V}^* \gets \arg \min_{\hat{V} \in \mathcal{V}} \widehat{MSE}_{\bbAutoOPE}\left(\hat{V}, \task^{(\text{new})}\right)$
\RETURN $\hat{V}^*$
\end{algorithmic}
\label{alg:auto-ope-selection}
\end{algorithm}

We recap the \bbAutoOPE method in Algorithms \ref{alg:auto-ope-training} (training of the model) and \ref{alg:auto-ope-selection} (zero-shot estimator selection). 
The \bbAutoOPE approach has several advantages. First of all, it is data-driven: \bbAutoOPE tries to learn the shared patterns in several different OPE tasks.
Furthermore, it is theory-inspired: we create a data-driven model inspired by the fact that there are some features of the OPE tasks which are theoretically proven to be related to the estimator's error. 
Finally, it is \textit{zero-shot}. This means that a practitioner can avoid training for each new estimator selection task, leading to a much faster computational time. 

A possible downside of our approach is that we trained \bbAutoOPE on a synthetic data set $\mathcal{M}$ created a priori, which may not lead to generalization to particular real-world OPE tasks.
In particular, Theorem \ref{thm:erm-bound}, provides the guarantee that our method generalizes on the same distribution of tasks. This implicitly assumes that, at test time, the OPE tasks come from the same distribution of the synthetic training tasks. However, this assumption might not hold. For this reason, we relax this assumption in the following:

\begin{theorem}\label{thm:domain}
Let $P_S$ and $P_T$ be the distributions of synthetic and real (test) tasks respectively. Let $L_S(f)$ and $L_T(f)$ be the expected loss of $f$ with respect to $P_S$ and $P_T$ respectively. Under the same assumptions of Theorem \ref{thm:erm-bound}, for any $\delta > 0$, with probability at least $1 - \delta$, we can bound 
$L_T$ of the empirical risk minimizer $\hat{f}$ as follows:
\[
L_{T}(\hat{f}) \leq  \sqrt{ \frac{8 \sigma^2 \log\left( \dfrac{2 |\mathcal{F}|}{\delta} \right)}{N} } + K \sqrt{d(P_S, P_T)}
\]
where $d$ is the chi-squared divergence, defined as $d(P_S, P_T) = \E_{P_S} [(( P_T(x) - P_S(x) ) / P_S(x))^2]$, and $K$ is a constant.

\end{theorem}

Also this finding is intuitive: we have that our approach learns to generalize on the test OPE tasks as long as it is trained on a large number of synthetic tasks and the divergence $d(P_S, P_T)$ is small. Notice that, the real OPE task distribution is unknown, and so is $d(P_S, P_T)$. 
In order to find whether $d(P_S, P_T)$ is small in practice or not, in Section \ref{sec:exp}, we test \bbAutoOPE on unseen real-world OPE tasks with an extensive experimental evaluation\footnote{We test on almost 50 different real-world experimental configurations, while prior work is usually tested on 2 or 3.} on a large number of real-world OPE tasks, obtained from different real datasets. 

\section{EXPERIMENTS}
\label{sec:exp}
In this section, we describe the experimental analysis of our proposed \bbAutoOPE and show its strong performance compared the \pasif baseline on
several real-world tasks, while exhibiting a lower computational cost. 

\subsection{
Experimental Settings}
\label{subsec:exp-common}

\paragraph{\pasif Baseline}
\label{par:synt-methods}
Our proposed method is compared to the \pasif baseline, which has specific hyperparameters. 
One important drawback of \pasif is the high computational cost, see Section \ref{par:computational_time}, which makes hyperparameter tuning impractical.\footnote{We estimate that applying the same Bayesian Search used for \bbAutoOPE would require at least a month of high-end GPU usage.
} For this reason we rely on the ad-hoc strategy presented in \citep{saito23} which tunes the regularization parameter $\lambda$ for each new OPE task via grid search. 
\vspace{-.5em}
\paragraph{\bbAutoOPE} We train the Random Forest model of \bbAutoOPE on the synthetic dataset $\mathcal{M}$ and we optimize its hyperparameters with a Bayesian search. For more details, we refer to Appendix \ref{app:bb-model}. In this Section, we only use the trained Random Forest in inference, in a \textit{zero-shot} way (see Algorithm \ref{alg:auto-ope-selection}) on new unseen tasks.

\vspace{-.5em}
\paragraph{Evaluation Metrics for OPE Selection}
\label{par:synt-metrics}
To evaluate \bbAutoOPE and \pasif on the Estimator Selection task we use two metrics: Relative Regret and Spearman's Rank Correlation Coefficient, as in \citep{saito23}, leveraging the fact that in all experiments we have access to the \textit{ground-truth} MSE of each estimator. Further details are in Appendix \ref{app:mse-gt}.

\textbf{Relative Regret} quantifies the relative difference between the MSE of the estimator identified by the Estimator Selection method, $\hat{V}_{\hat{m}_y},\ y \in \{\pasif, \bbAutoOPE\}$, and the MSE of the best estimator, $\hat{V}_{m^*}$. This difference is then normalized by the ground-truth MSE of the best estimator:
\begin{align*}
    & \text{R-Regret} \coloneqq \frac{\text{MSE}(\hat{V}_{\hat{m}_y}(\pi_e)) - \text{MSE}(\hat{V}_{m^*}(\pi_e))}{\text{MSE}(\hat{V}_{m^*}(\pi_e))} \ .
\end{align*}

\textbf{Spearman's Rank Correlation Coefficient} \citep{spearman1961proof} measures the similarity between two rankings, one based on the MSE predicted by the Estimator Selection method and the other based on the ground-truth MSE. 

\subsection{Results}
\label{subsec:real-exp}


\subsubsection{Open Bandit Dataset (OBD)}
\label{subsubsec:obd}
In this real-world experiment, we use the Open Bandit Dataset (OBD) \citep{obp}, a publicly available dataset originated from a large-scale fashion e-commerce platform. The dataset encompasses three campaigns: `ALL', `MEN', and `WOMEN'. In each campaign, the deployed policy is randomly chosen at each round, selecting either the Uniform Random policy or the Bernoulli Thompson Sampling policy.
\vspace{-.5em}
\paragraph{CB Data}
\label{par:obd-data}
We focus on the `ALL' campaign, and opt to employ the bandit feedback collected under the Uniform Random policy as logging data, while the data collected under the Bernoulli Thompson Sampling policy serves as evaluation data.
In order to compute confidence intervals, we perform stratified bootstrap sampling 20 times.
Each subsample consists of around 35,000 rounds
 and constitutes a new OPE task.
\vspace{-.5em}
\paragraph{Results}
\label{par:obd-res}

\begin{table}
\centering
\small
\begin{tabular}{l|l|l}
\toprule 
 Method &  Rel. Regret ($\downarrow$) & Spearman ($\uparrow$)\\
\midrule
 \pasif & 4.54 $\pm$ 4.47 & 0.08 $\pm$ 0.46  \\
\hline
 \bbAutoOPE & \bfseries 0.79 $\pm \approx$ 0.00 & \bfseries 0.69 $\pm$ 0.05 \\
\bottomrule
\end{tabular}
\caption{Experimental results on Open Bandit Dataset (OBD).}
\vspace{-1em}
\label{tab:obd-res}
\end{table}
The outcomes of this experiment are shown in Table \ref{tab:obd-res}. Our \bbAutoOPE method exhibits a Relative Regret almost 6 times lower compared to \pasif, and the Spearman's Rank Correlation Coefficient is much better than \pasif. \bbAutoOPE confirms its stability in its predictions, exhibiting low variance, while \pasif has a large variance especially on Relative Regret.
This experiment confirms that \bbAutoOPE, while being trained on synthetic data, is able to generalize on data coming from real distributions.

\subsubsection{UCI datasets}
\label{subsubsec:uci}
To further strengthen our empirical evaluation of the generalization capabilities of \bbAutoOPE, in this section we present 40 other real-world experiments conducted on 8 different datasets. In particular, we took 8 datasets from the UCI repository \citep{uci}, as typically done in the OPE literature 
\citep{beygelzimer2009offset, dros, dr3, robustness-off-policy}. We report an overview of these datasets in Table \ref{tab:uci-characteristics}.

\begin{table*}
\small
    \centering
    $$
    \begin{array}{l|llllllll}
    \hline
    \text{Dataset} & \text{letter} & \text{optdigits} & \text{page-blocks} & \text{pendigits} & \text{satimage} & \text{vehicle} & \text{yeast} & \text{breast-cancer} \\
    \hline
    \text{Classes} & 26 & 10 & 5 & 10 & 6 & 4 & 10 & 2 \\
    \text{Sample size} & 20000 & 5620 & 5473 & 10992 & 6435 & 846 & 1484 & 569 \\
    \hline
    \end{array}
    $$
    \caption{Characteristics of the considered UCI datasets.}
    \label{tab:uci-characteristics}
\end{table*}

\paragraph{CB Data}
\label{par:uci-data}
To adapt these datasets for CB we perform the standard supervised to bandit conversion \citep{beygelzimer2009offset, joachims2018deep}, reinterpreting the input features as context features and the class labels as actions, while the reward is 1 if the current action is the correct class for the given context, 0 vice versa.
Furthermore, we partition the dataset into two subsets, the first one is used as logging data $\mathcal{D}_b$ for the OPE task, while the second is used to generate both logging and evaluation policies. In order to generate the policies, the second data split is used to train two Logistic Regression classifiers. Each classifier defines a deterministic policy $\pi_{det}$ because, given a context, it predicts a specific class label (action) with a probability of 1, while all other labels having a probability of 0. 
To produce the final stochastic policies we
blend the deterministic policy with a uniform random policy $\pi_u(a|x)$ using parameters $\alpha_b$, $\alpha_e$:
\begin{align*}
     \pi_b(a|x) \coloneqq \alpha_b \cdot \pi_{det, b}(a|x) + (1 - \alpha_b) \cdot \pi_u(a|x) \ \ \ , \\
     \pi_e(a|x) \coloneqq \alpha_e \cdot \pi_{det, e}(a|x) + (1 - \alpha_e) \cdot \pi_u(a|x) \ \ \ ,
\end{align*}
with $0 \leq \alpha_e \leq 1 \ , \ \ 0 \leq \alpha_b \leq 1 $. We set $\alpha_b = 0.2$ for the logging policy and $\alpha_e \in \left\{0, 0.25, 0.5, 0.75, 0.99\right\}$ to create diverse evaluation policies, generating different OPE tasks. In the end, we generate 40 different OPE tasks (8 datasets, 5 different $\alpha_e$ values). In order to compute the confidence intervals, we perform 50 stratified bootstrap samples of 90\% of the logging data.


\paragraph{Results}

{\renewcommand{\arraystretch}{-0}
\begin{figure*}
\begin{tabular}{cc}
\subfloat{\includegraphics[width=.98\columnwidth]{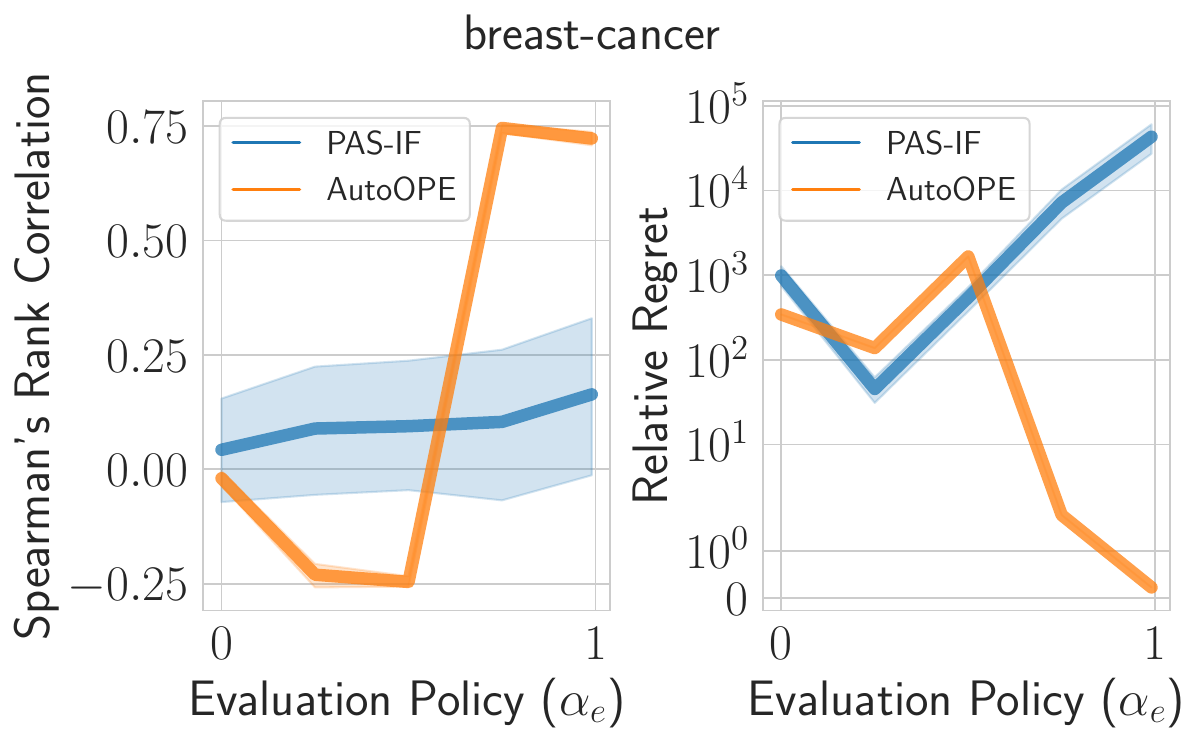}} &
\subfloat{\includegraphics[width=.98\columnwidth]{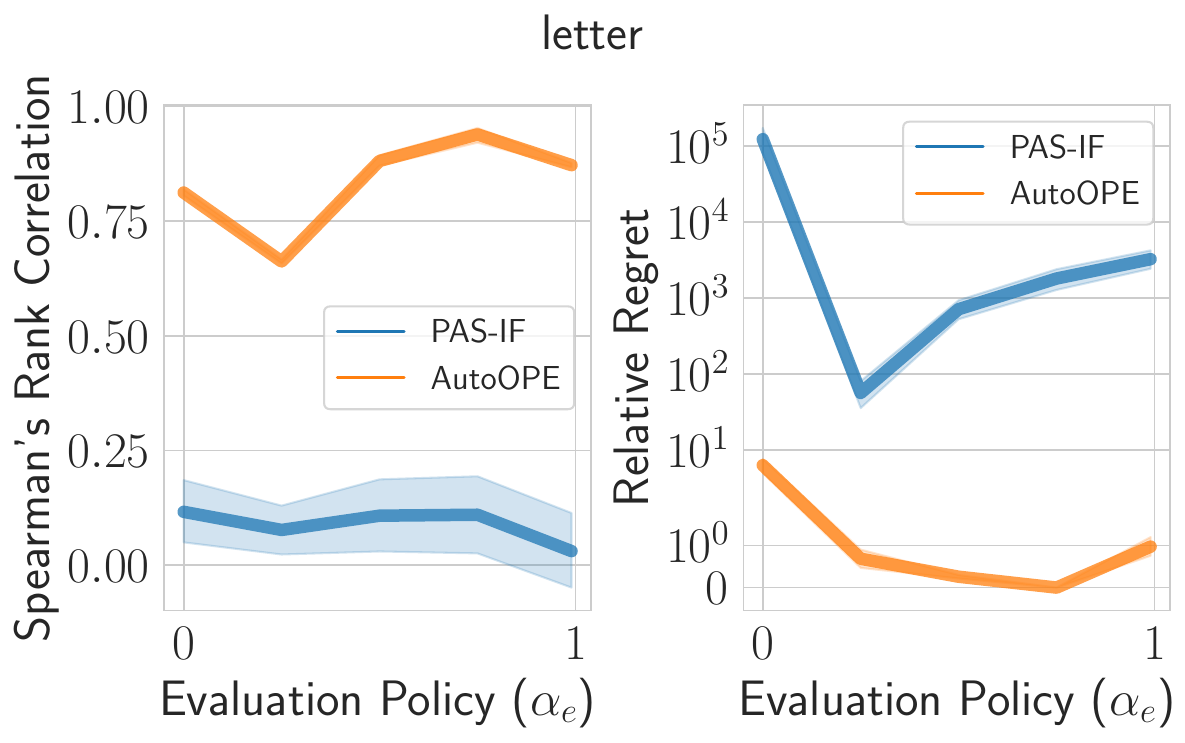}} \\
\subfloat{\includegraphics[width=.98\columnwidth]{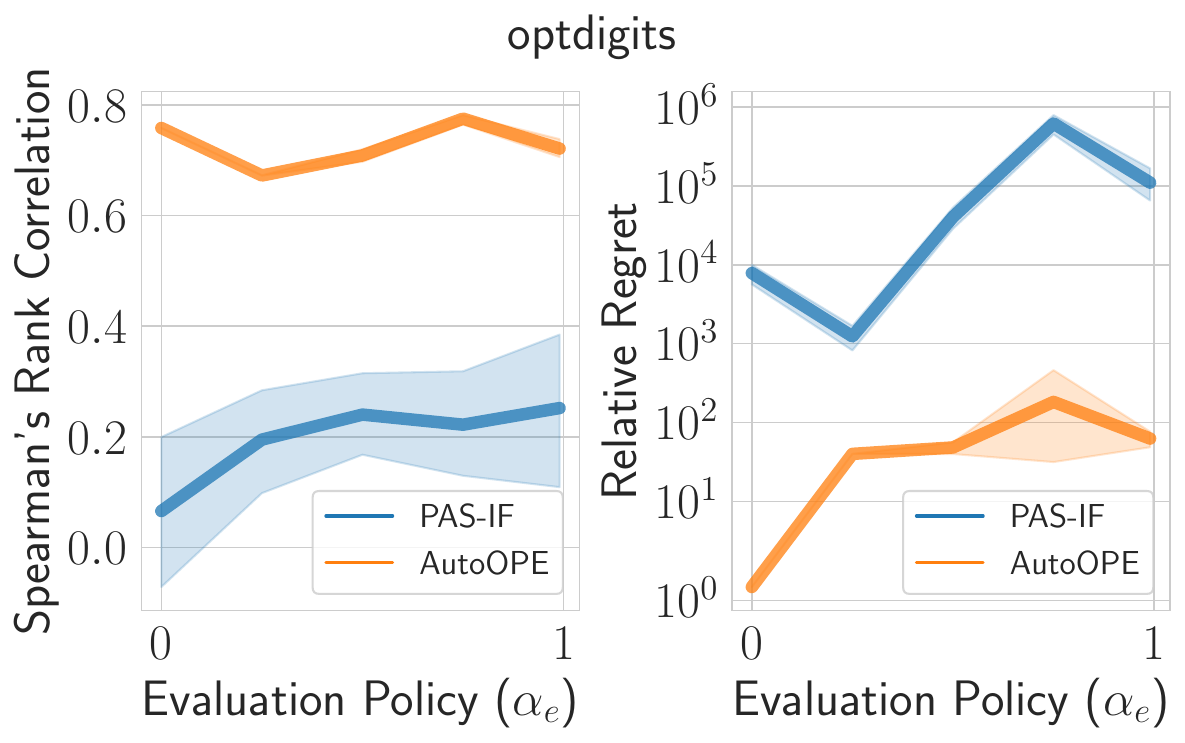}} &
\subfloat{\includegraphics[width=.98\columnwidth]{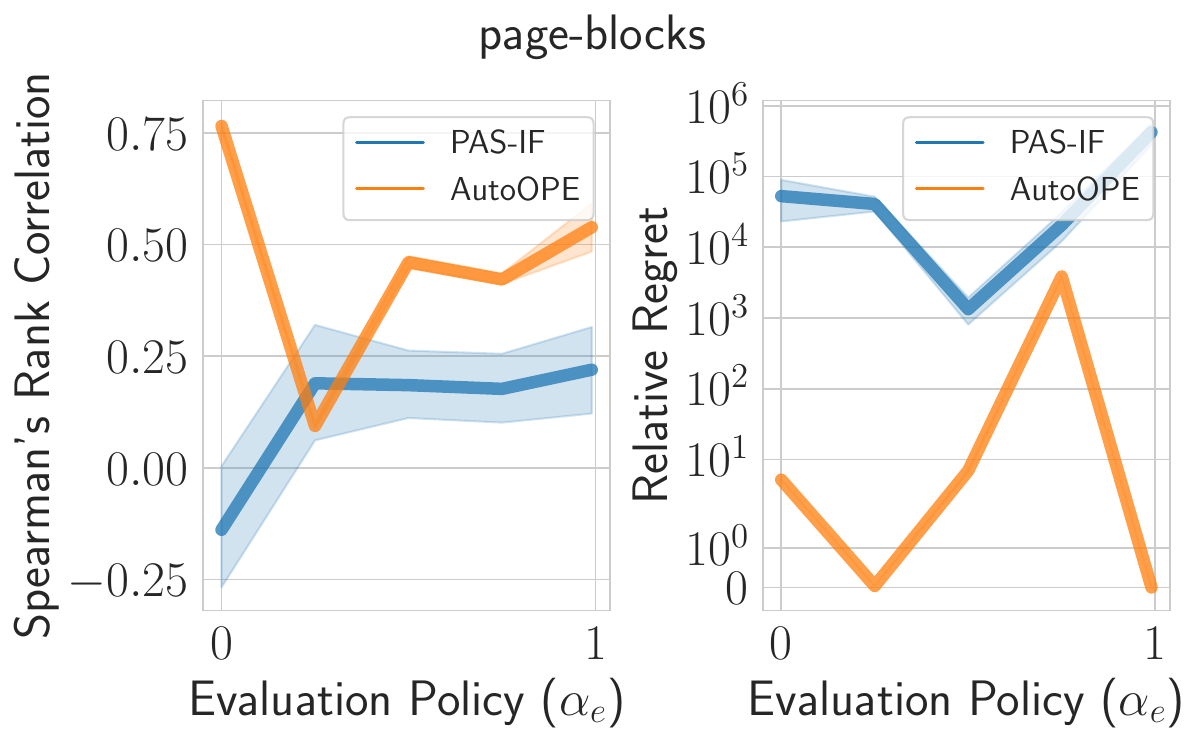}} \\
\subfloat{\includegraphics[width=.98\columnwidth]{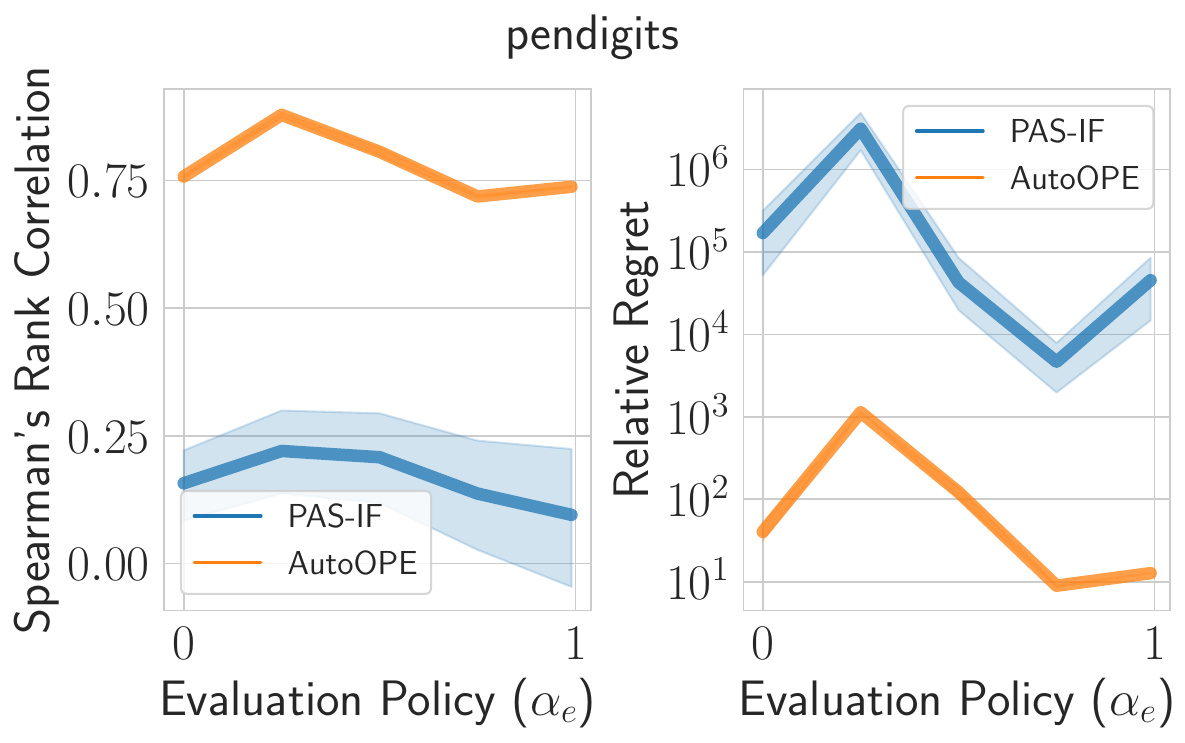}} &
\subfloat{\includegraphics[width=.98\columnwidth]{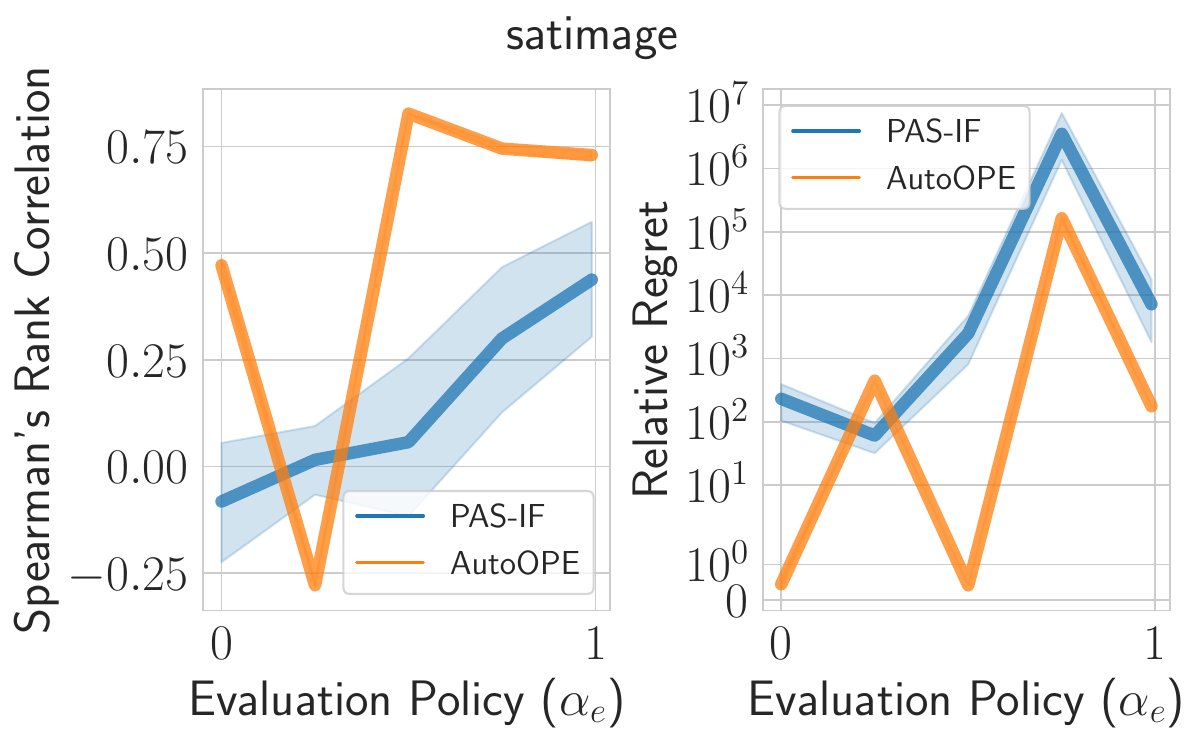}} \\
\subfloat{\includegraphics[width=.98\columnwidth]{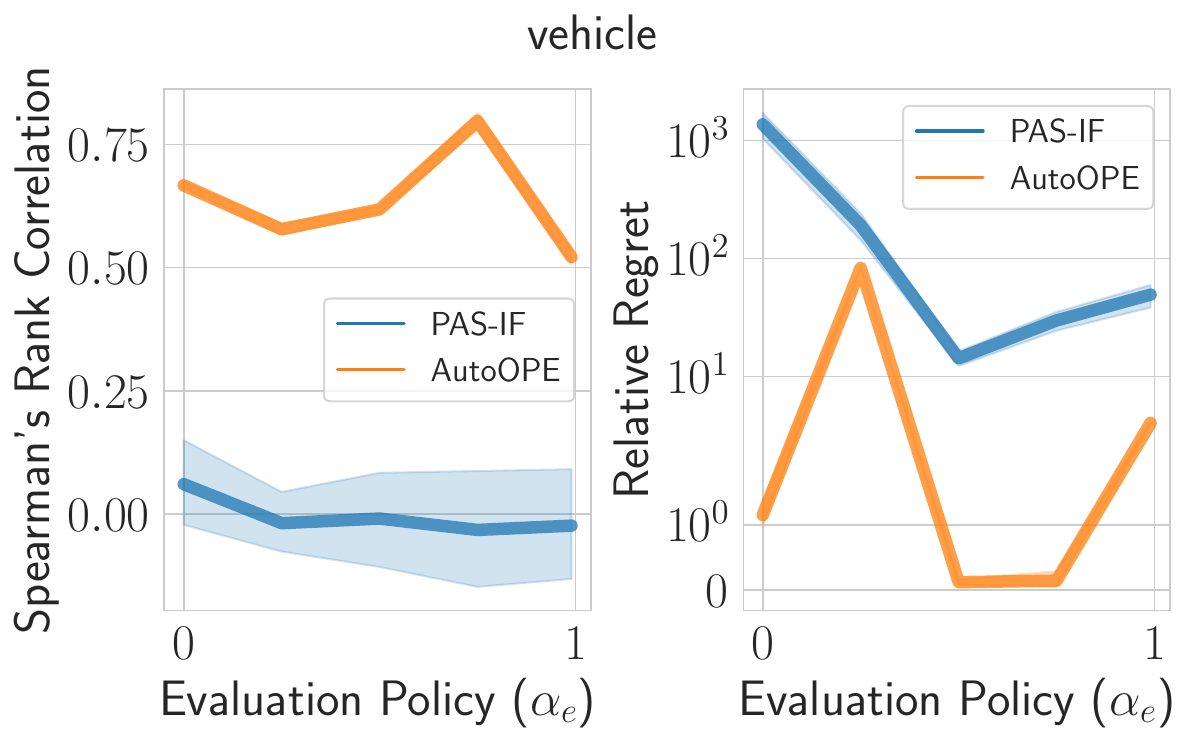}} &
\subfloat{\includegraphics[width=.98\columnwidth]{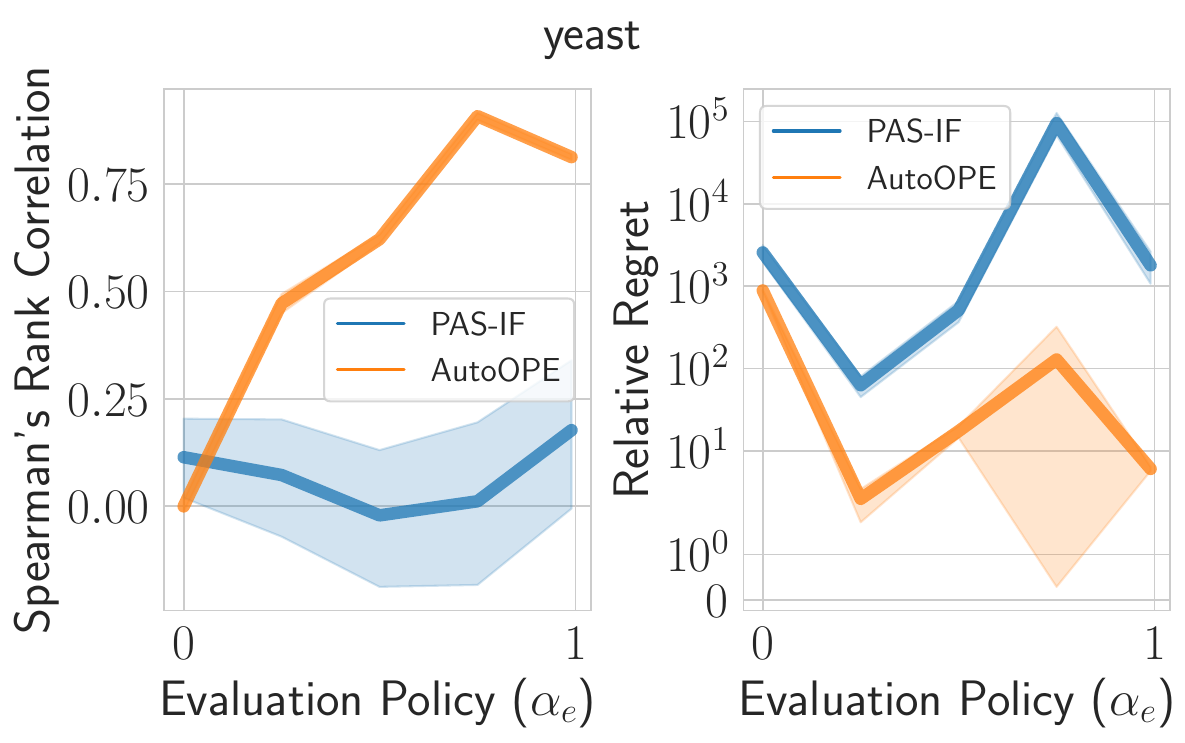}}
\end{tabular}
\caption{Relative Regret (lower is better) and Spearman's Rank Correlation Coefficient (higher is better) for Estimator Selection experiments on various UCI datasets (the name of the used dataset is in the title of the corresponding plot). Shaded areas correspond to 95\% confidence intervals.}
\label{fig:uci-res}
\end{figure*}}



The results are shown in Figure \ref{fig:uci-res}. \bbAutoOPE consistently outperforms \pasif, demonstrating lower Relative Regret and better Spearman's Rank Correlation Coefficient in almost all the 40 different experimental configurations. Indeed, \bbAutoOPE outperforms \pasif 36 out of 40 configurations in terms of Spearman's Coefficient, and 37 out of 40 configurations in terms of Relative Regret, even if \bbAutoOPE performs estimator selection in a zero-shot way. Notably, when \bbAutoOPE outperforms \pasif, it does it by a large margin.  Also, \bbAutoOPE exhibits lower variance in particular for the Spearman's coefficient, while \pasif exhibits very high variance. This comprehensive empirical evaluation and its results are consistent with the OBD experiments and confirm that \bbAutoOPE is able to generalize to unseen OPE tasks with a different distribution compared to the one used for its training, in a zero-shot fashion.

\subsection{Computational Time Comparison}
\label{par:computational_time}
\begin{figure*}
\begin{minipage}{0.99\columnwidth}
    \includegraphics[width=\columnwidth]{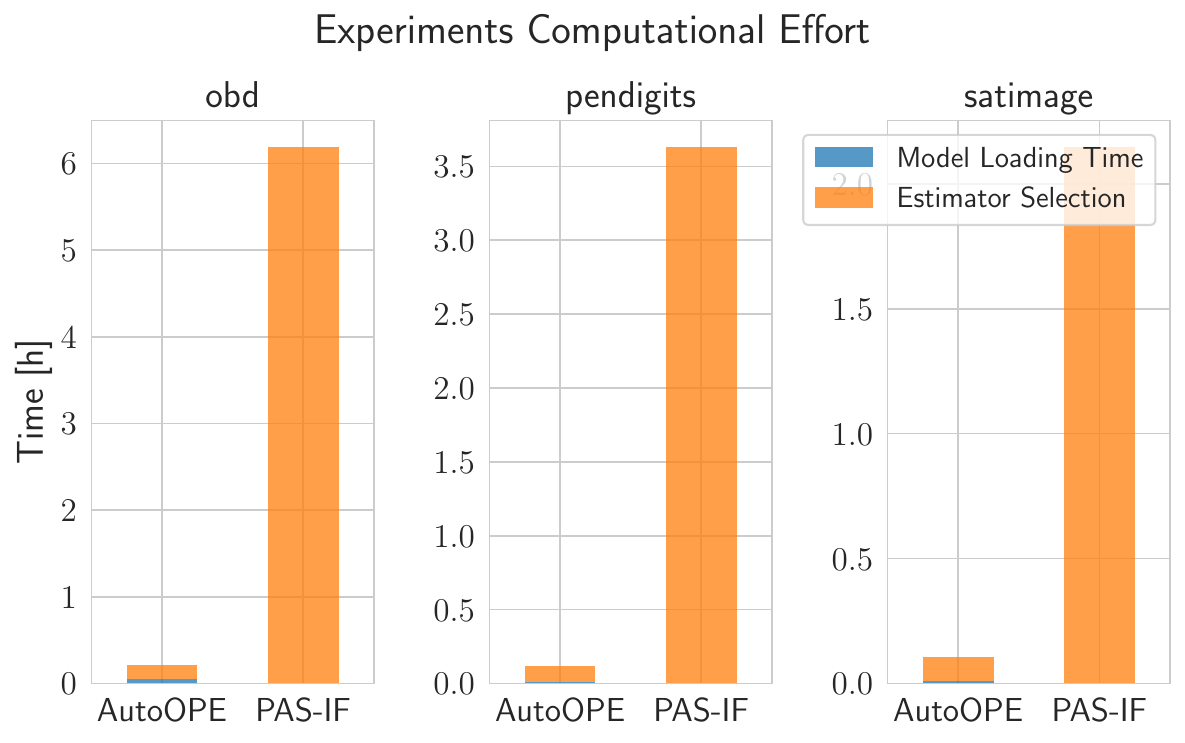}
    \caption{Comparison on the computational cost of \bbAutoOPE and \pasif for the Estimator Selection task.
    }
    \label{fig:comp-time-exp3}
    
\end{minipage}
\hfill
\begin{minipage}{0.99\columnwidth}
\includegraphics[width=\columnwidth]{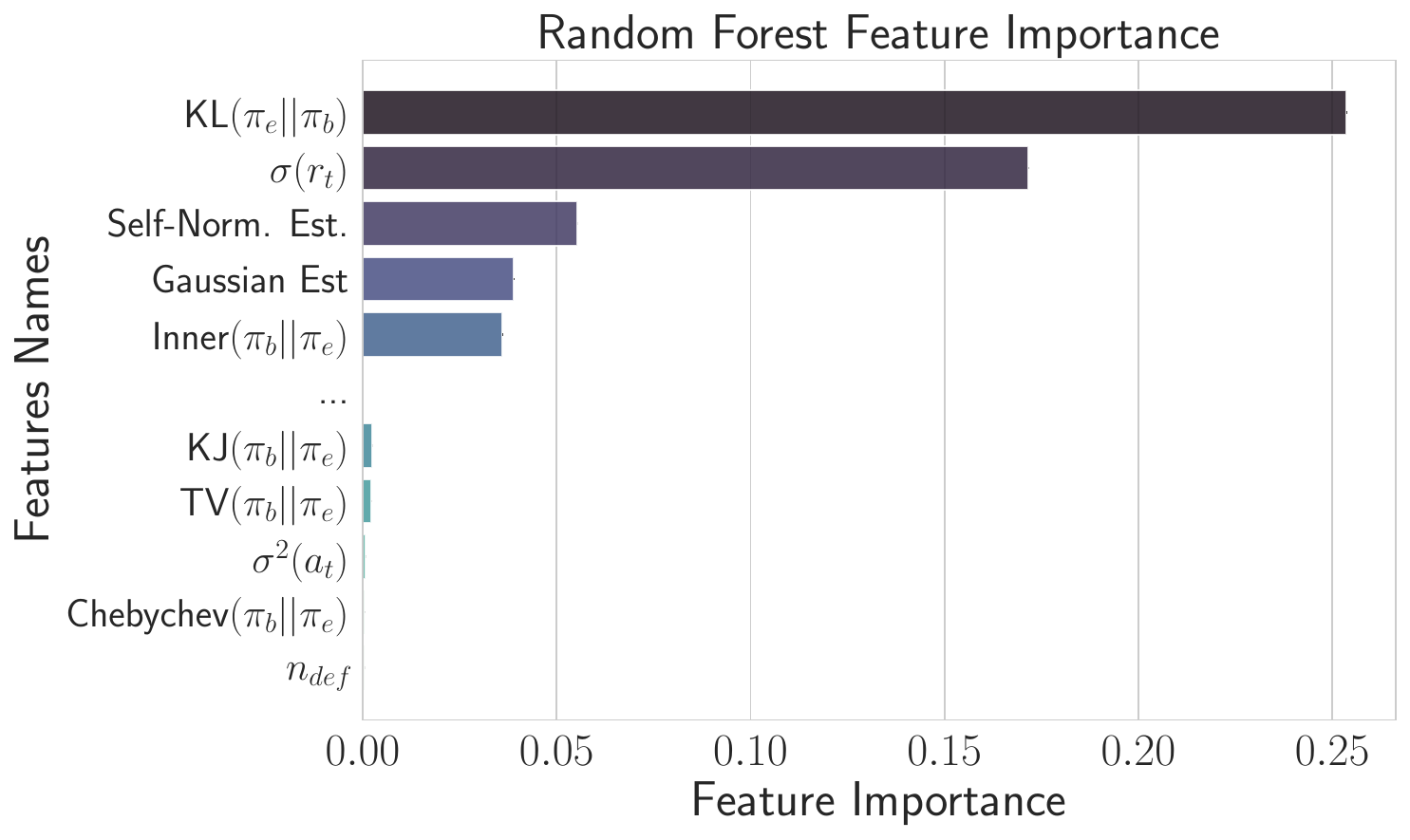}
    \caption{Feature importance of the most and least important features for 
    \bbAutoOPE.}
    \label{fig:preview-feature-importance}
\end{minipage}
\end{figure*}

One important advantage of our \bbAutoOPE method is that, after the pre-training phase, it is ready-to-use for inference, potentially on many different datasets and evaluation policies. On the other hand, the baseline method \pasif must be trained from scratch and perform hyperparameter optimization for every new evaluation policy and dataset. Due to this, the time required to perform the Estimator Selection task is very different.
Figure \ref{fig:comp-time-exp3} reports a comparison of the time required to perform the Estimator Selection task.
As expected, \bbAutoOPE is in the order of 10 times faster compared to \pasif, even though \pasif performs estimator selection on a GPU, while \bbAutoOPE does it on a CPU. Further details are reported in Appendix \ref{app:infrastructure}.

\subsection{Additional Experiments}

Figure \ref{fig:preview-feature-importance} shows a selection of features for the Random Forest model. The full plot is in Figure \ref{fig:full-feature-importance}, Appendix \ref{app:feat-eng}, together with its discussion. The insights gained from this feature importance analysis can shed light on future theoretical findings, lead by analyzing the features that had a significant impact on the model's performance.
Also, we conducted additional experiments on estimator selection on synthetic datasets and on the CIFAR dataset \citep{cifar} reported in Appendix \ref{app:additional-exp}. 
Finally, we conducted different ablation studies, investigating the impact of scaling the size of the synthetic dataset, the diversity of the dataset, and the features used. We describe all these ablation experiments in Appendix \ref{app:ablation}.

\section{CONCLUSIONS}
\label{sec:conclusions}
In this paper, we focused on the Estimator Selection problem in Off-Policy Evaluation. We showed that, despite its significance, it is an overlooked problem in the literature. We proposed \bbAutoOPE, a novel data-driven approach for Estimator Selection, which leverages synthetic OPE tasks in order to learn to generalize the estimator selection capabilities to unseen real-world OPE tasks. Through several experiments, we empirically showed how our method is able to consistently make a better estimator selection when compared to the baseline method on several real-world datasets, outperforming the baseline method by a large margin with a significantly lower computational cost. 

For future work, a natural direction to follow would be to try different machine learning models for \bbAutoOPE.
Furthermore, one could try to theoretically analyze the relationship between the features of the \bbAutoOPE dataset and the performance of the various estimators, which was unveiled by the feature importance of the random forest model. 
Finally, our approach focused on the estimator selection problem in the contextual bandit domain. Our method could be used as a stepping stone towards OPE estimator selection in the more general Reinforcement Learning scenario.




\clearpage

\bibliographystyle{plainnat}
\bibliography{biblio}

\clearpage
\appendix
\onecolumn

\section{RELATED WORK}
\label{app:rel-work}
In the realm of Contextual Bandit Off-Policy Evaluation, there exists a rich and diverse landscape of published research aimed at assessing the performance of policies without actually deploying them, avoiding the hazard of selecting action under not valuable or deleterious policies which, depending on the scenario, could be costly and dangerous \citep{gilotte, levine, saito-johachins-21, kk-saito-21}. Indeed, this area of study has brought contributions to various application domains, including recommender systems \citep{DBLP:conf/www/LiCKG15}, advertising \citep{DBLP:journals/jmlr/BottouPCCCPRSS13}, and precision medicine \citep{medicine}, reflecting its flexibility in modeling different situations. In this context, \bbAutoOPE emerges as a novel approach, offering a unique solution based on a supervised learning model that leverages synthetic data and supervised learning to tackle the Estimator Selection problem. 

To appreciate the significance of \bbAutoOPE, it is essential to understand the broader context of Contextual Bandit OPE research.

Several pioneering works have laid the foundation for evaluating counterfactual policies in the context of Contextual Bandit. Notable contributions include studies that have introduced fundamental approaches such as the Direct Method (DM) \citep{dm}, Inverse Propensity Scoring (IPS) \citep{ips1, ips2}, and Doubly Robust (DR) \citep{dr1, dr2, dr3}, which serve as the cornerstone for OPE research, since they define the three main classes of OPE estimators: model-based, model-free and hybrid.

Direct Method (DM) is rooted in machine learning, employing algorithms to estimate policy performance by regressing rewards. Its effectiveness depends on the precision of the reward estimation produced by the learning algorithm. For this reason, its performance may decrease specially in complex environments like industrial recommender systems. DM is susceptible to bias when the model does not produce accurate reward estimations, caused by the inability to correct the distribution shift between the observed data and the counterfactual policy. 
It is a purely model-based estimator. 

Inverse Propensity Scoring (IPS) instead is purely model-free. IPS is unbiased on expectation, thanks to the employment of the notion of importance sampling by statistics, but exhibits high variance, that increases when logging and evaluation policies are significantly different. 

Doubly Robust (DR) represents a blend of DM and IPS characteristics, since it is unbiased and generally presents a lower variance compared to IPS. However, its performance still depends on the specific task so it is not necessarily the best option.

A multitude of OPE estimators, has been designed over time as extensions or combinations of DM, IPS and DR \citep{snips1, snips2, dros, subgaus, dr1, switch}. These efforts collectively contribute to advance OPE methodologies.

However, many estimators incorporate hyperparameters to balance the bias-variance trade-off given the Contextual Bandit environment such as \citep{dros, subgaus, dr1, dr2, dr3, switch}. As reported by \citep{robustness-off-policy, obp} the hyperparameter tuning strongly affects the prediction quality. To address the issue of hyperparameter optimization 
\citep{slope} and \citep{slope++} proposed SLOPE and its improved version SLOPE++, based on Lepski's principle \citep{lepski}, offering improved results compared to na\"{\i}ve tuning procedures. Lepski's principle allows to define an upper-bound for the Mean Squared Error without delving into the bias estimation, that is the main reason why the task of MSE estimation is challenging.
SLOPE dominates all the other methods designed for hyperparameter tuning, being the state of the art technique for this task. Other existing methods like MAGIC \citep{dr1}, or other built-in methods of proposed estimators \citep{dros, switch} have tackled hyperparameter selection using MSE estimators, but they offer weaker guarantees compared to SLOPE.

The Estimator Selection problem has emerged as a relatively unexplored but pivotal aspect of OPE research. Empirical studies by \citep{voloshin} and \citep{robustness-off-policy} have shown the fundamental role of Estimator Selection on OPE accuracy, highlighting its importance in practical applications. 

In this context, the recent \pasif (Policy-Adaptive Estimator Selection via Importance Fitting) method \citep{saito23} emerges as the only baseline  for Estimator Selection. \pasif uses a neural network to learn a subsampling rule that is used to generate two pseudo-datasets starting from the available logging data. The subsampling rule is learned by optimizing an importance fitting loss that depends on the specific logging and evaluation policies one wishes to evaluate. Once the two pseudo datasets have been generated, \pasif allows to estimate the candidate estimators' Mean Squared Error based on the empirical average of rewards in these pseudo datasets.


Our proposed \bbAutoOPE method, on the other hand, competes with \pasif in performance by introducing a novel data-driven approach that leverages learning from synthetic data with a supervised model. \bbAutoOPE offers a versatile approach, allowing the comparison of different estimator classes and the selection of the most accurate one. This innovative approach empirically promises to significantly enhance the accuracy of OPE, reducing an important gap in the Contextual Bandit Off-Policy Evaluation field.

\section{DEFERRED PROOFS}
\label{app:proofs}

\setcounter{theorem}{0}

\begin{theorem}
Under Assumptions \ref{ass:mse-func}, \ref{ass:subgaussian-loss}, and \ref{ass:finite-hyp}, suppose we have $N = |\mathcal{V}| |\mathcal{M}|$ independent samples $\{(x_i, y_i)\}_{i=1}^N$, where each $x_i = g(\hat{V}_i, \mathcal{T}_i)$ and $y_i = \text{MSE}(\hat{V}_i(\pi_e); \mathcal{T}_i)$, drawn uniformly over the estimators $\hat{V} \in \mathcal{V}$ and tasks $\mathcal{T} \in \mathcal{M}$. Let $\hat{f}$ be the empirical risk minimizer over the finite hypothesis class $\mathcal{F}$:
\[
\hat{f} = \arg \min_{f \in \mathcal{F}} \frac{1}{N} \sum_{i=1}^N \mathcal{L}\left( y_i, f(x_i) \right).
\]
Then, for any $\delta > 0$, with probability at least $1 - \delta$, the expected loss $L(\hat{f})$ satisfies:
\[
L(\hat{f}) \leq  \sqrt{ \frac{8 \sigma^2 \log\left( \dfrac{2 |\mathcal{F}|}{\delta} \right)}{N} },
\]
where $\sigma^2$ is the subgaussian parameter from Assumption \ref{ass:subgaussian-loss}, and the expected loss $L(f)$ is defined as
\[
L(f) = \mathbb{E}_{(x, y)} \left[ \mathcal{L}\left( y, f(x) \right) \right].
\]
\end{theorem}

\begin{proof}
Under Assumption \ref{ass:mse-func}, there exists a function $f^* \in \mathcal{F}$ such that for all $(\hat{V}, \mathcal{T})$,
\[
y = \text{MSE}(\hat{V}(\pi_e); \mathcal{T}) = f^*(x) = f^*(g(\hat{V}, \mathcal{T})).
\]
This implies that the expected loss of $f^*$ is zero:
\[
L(f^*) = \mathbb{E}_{(x, y)} \left[ \mathcal{L}\left( y, f^*(x) \right) \right] = 0.
\]

Let $\Delta(f) = L_\mathcal{M}(f) - L(f)$ denote the deviation between the empirical and expected losses, where
\[
L_\mathcal{M}(f) = \frac{1}{N} \sum_{i=1}^N \mathcal{L}\left( y_i, f(x_i) \right).
\]
By Assumption \ref{ass:subgaussian-loss}, for any fixed $f \in \mathcal{F}$, the random variable $\mathcal{L}\left( y_i, f(x_i) \right)$ is subgaussian with parameter $\sigma^2$. Using Hoeffding's inequality for subgaussian variables, we have for any $\epsilon > 0$:
\[
\mathbb{P}\left( |\Delta(f)| \geq \epsilon \right) \leq 2 \exp\left( -\frac{N \epsilon^2}{2 \sigma^2} \right).
\]
Since $\mathcal{F}$ is finite, applying the union bound over all hypotheses:
\[
\mathbb{P}\left( \sup_{f \in \mathcal{F}} |\Delta(f)| \geq \epsilon \right) \leq 2 |\mathcal{F}| \exp\left( -\frac{N \epsilon^2}{2 \sigma^2} \right).
\]
Setting the right-hand side equal to $\delta$ and solving for $\epsilon$, we get:
\[
\epsilon = \sigma \sqrt{ \frac{2 \log\left( \dfrac{2 |\mathcal{F}|}{\delta} \right)}{N} }.
\]

On the event where $\sup_{f \in \mathcal{F}} |\Delta(f)| \leq \epsilon$, the following holds for the empirical risk minimizer $\hat{f}$:
\[
L(\hat{f}) = L_\mathcal{M}(\hat{f}) - \Delta(\hat{f}) \leq L_\mathcal{M}(\hat{f}) + \epsilon.
\]
Since $L_\mathcal{M}(\hat{f}) \leq L_\mathcal{M}(f^*)$ by definition of $\hat{f}$, and $L_\mathcal{M}(f^*) = L(f^*) + \Delta(f^*) = 0 + \Delta(f^*) \leq \epsilon$, we have:
\[
L(\hat{f}) \leq L_\mathcal{M}(\hat{f}) + \epsilon \leq L_\mathcal{M}(f^*) + \epsilon \leq \epsilon + \epsilon = 2 \epsilon.
\]
Substituting $\epsilon$ yields the desired bound:
\[
L(\hat{f}) \leq 2 \sigma \sqrt{ \frac{2 \log\left( \dfrac{2 |\mathcal{F}|}{\delta} \right)}{N} } = \sqrt{ \frac{8 \sigma^2 \log\left( \dfrac{2 |\mathcal{F}|}{\delta} \right)}{N} }.
\]
\end{proof}




\begin{theorem}
Let $P_S$ and $P_T$ be the distribution of synthetic and real test tasks respectively. Let $L_S(f)$ and $L_T(f)$ be the expected loss of $f$ with respect to $P_S$ and $P_T$ respectively. Under the same assumptions of Theorem \ref{thm:erm-bound}, for any $\delta > 0$, with probability at least $1 - \delta$, we can bound 
$L_T$ of the empirical risk minimizer $\hat{f}$ as follows:
\[
L_{T}(\hat{f}) \leq  \sqrt{ \frac{8 \sigma^2 \log\left( \dfrac{2 |\mathcal{F}|}{\delta} \right)}{N} } + K \sqrt{d(P_S, P_T)}
\]
where $d$ is a divergence defined as $d(P_S, P_T) = \E_{P_S} [(( P_T(x) - P_S(x) ) / P_S(x))^2]$, and $K$ is a constant.

\end{theorem}
\begin{proof}

For any $f \in \mathcal{F}$:
\begin{equation}\label{eq:div-bound}
    \begin{aligned}
        L_T(f) & = L_S(f) + L_T(f) - L_S(f) & \\
        & \leq L_S(f) + \left\lvert L_T(f) - L_S(f) \right\rvert & \eqnote{(apply absolute value)} \\ 
        & =  L_S(f) + \left\lvert \int_x \mathcal{L}(f(x), f^*(x)) \left( P_T(x) - P_S(x) \right) dx \right\rvert & \\
        & =  L_S(f) + \left\lvert \int_x \mathcal{L}(f(x), f^*(x)) \sqrt{P_S(x)} \frac{\left( P_T(x) - P_S(x) \right)}{\sqrt{P_S(x)}}  dx \right\rvert &  \eqnote{(multiply and divide by $\sqrt{P_S(x)}$)}\\
        & \leq L_S(f) + \sqrt{\int_x \mathcal{L}(f(x), f^*(x))^2 P_S(x) dx} \sqrt{\int_x \frac{\left( P_T(x) - P_S(x) \right)^2}{P_S(x)} dx } & \eqnote{(Cauchy–Schwarz)} \\ 
        & \leq L_S(f) + K \sqrt{\int_x \frac{\left( P_T(x) - P_S(x) \right)^2}{P_S(x)} dx } & \eqnote{(a subgaussian r.v. $Z$ has $\E[Z^2] \leq K^2$, with $K$ a constant)} \\
        & = L_S(f) + K \sqrt{d(P_S, P_T)} & 
    \end{aligned}
\end{equation}

where $d$ is the chi-squared divergence, defined as:
\begin{equation*}
 d(P_S, P_T) =  \int_x \frac{\left( P_T(x) - P_S(x) \right)^2}{P_S(x)} dx \ .
\end{equation*}

If we consider the empirical risk minimizer $\hat{f}$ on $\mathcal{M} \overset{\mathrm{iid}}{\sim} P_S$ defined as:
\[
\hat{f} = \arg \min_{f \in \mathcal{F}} \frac{1}{N} \sum_{(x_i, y_i) \in \mathcal{M}} \mathcal{L}\left( y_i, f(x_i) \right) ,
\]
and we apply Theorem \ref{thm:erm-bound}, we have that:
\begin{equation*}
    \begin{aligned}
        L_T(\hat{f}) & \leq L_S(\hat{f}) +  K \sqrt{d(P_S, P_T)} & \eqnote{(applying Eq. \ref{eq:div-bound})} \\
        & \leq  \sqrt{ \frac{8 \sigma^2 \log\left( \dfrac{2 |\mathcal{F}|}{\delta} \right)}{N} } + K \sqrt{d(P_S, P_T)} \ , & \eqnote{(applying Theorem \ref{thm:erm-bound})}
    \end{aligned}
\end{equation*}

with probability at least $1 - \delta$.

\end{proof}

\section{OFF-POLICY ESTIMATORS DESCRIPTION}
\label{app:ope-est}
This Appendix contains a brief characterization of all the OPE estimators, $\mathcal{V}$, used in this paper. Note that the total number of OPE candidate estimators is 21 which includes 3 model-free estimators and 6 requiring a reward predictors, of which three were included. 

\subsection{Model-Based Estimators}
We include one estimator based exclusively on a reward model.

\paragraph{Direct Method (DM)}
The DM estimator \citep{dm} employs a supervised machine learning model $\hat{q}$ to estimate the mean reward $\mathbb{E}_{p(r_t|x_t, a_t)}\left[r_t\right]$. The policy value prediction is computed in this way:
\begin{align*}
    \hat{V}_{DM}(\pi_e; \task, \hat{q}) \coloneqq \frac{1}{n_b} \sum_{t}^{n_b} \sum_{a \in \mathcal{A}} \pi_e(a|x_t) \hat{q}(x_t, a) \ \ .
\end{align*}
It is a method that exhibits a small variance in general, but since its performance is strictly dependent on the reward model precision, the bias can be high when the reward estimation is not accurate. It can be a good choice when the complexity of the task is not too high.

\subsection{Model-Free Estimators}
We include three estimators that do not rely on a reward predictor, but instead are built on the statistical concept of \textit{importance sampling}.

\paragraph{Inverse Propensity Scoring (IPS)}
The IPS estimator \citep{ips1, ips2} employs pure importance sampling to re-weight the average of rewards, accounting for the distribution shift between $\pi_b$ and $\pi_e$, since the average of rewards are collected under the logging policy, while the goal is to evaluate $\pi_e$.
The policy value prediction is computed in this way:
\begin{align*}
    \hat{V}_{IPS}(\pi_e; \task) \coloneqq \frac{1}{n_b} \sum_{t}^{n_b} w(x_t, a_t) r_t,\ \ \ \ w(x_t, a_t) \coloneqq \frac{\pi_e(a_t|x_t)}{\pi_b(a_t|x_t)} \ \ .
\end{align*}
IPS is provably unbiased, if the condition of full support holds, and consistent, but it usually exhibits high variance, especially with small sample data or with divergent $\pi_b,\ \pi_e$ (in this case the importance weight $w$ could blow up, leading to instability).

\paragraph{Self-Normalized Inverse Propensity Scoring (SNIPS)}
The SNIPS estimator \citep{snips1, snips2} tries to stabilize the IPS estimator normalizing the prediction by the empirical mean of the importance weights.
The policy value prediction is computed in this way:
\begin{align*}
    \hat{V}_{SNIPS}(\pi_e; \task) \coloneqq \frac{\sum_{t}^{n_b} w(x_t, a_t) r_t}{\sum_{t}^{n_b} w(x_t, a_t)} \ \ .
\end{align*}
It effectively reduces the variance in many situations with respect to IPS, and it is still consistent. However, it is biased.

\paragraph{Subgaussian Importance Sampling Inverse Propensity Scoring (IPS-$\lambda$)}
The IPS-$\lambda$ estimator \citep{subgaus} shrink the importance ratio smoothly, with a sub-gaussian parameterized by a value $\lambda$ to reduce the IPS variance. The policy value prediction is computed in this way:
\begin{align*}
    & \hat{V}_{IPS-\lambda}(\pi_e; \task, \lambda, \gamma) \coloneqq \frac{1}{n_b} \sum_{t}^{n_b} w_{\lambda, \gamma}(x_t, a_t) r_t,
    \\ &  w_{\lambda, \gamma}(x_t, a_t) \coloneqq ((1 - \lambda) w(x_t,a_t)^\gamma + \lambda)^{\frac{1}{\gamma}},\ \ \lambda \in \left[0,1\right],\ \gamma \leq 1 \ \ .
\end{align*}
The $\lambda,\ \gamma$ hyperparameters balance the bias-variance trade-off of the estimator.

\subsection{Hybrid Estimators}
We include five hybrid estimators, that leverage both a reward predictor and importance sampling.

\paragraph{Doubly Robust (DR)}
The DR estimator \citep{dr1, dr2, dr3} employs a reward model $\hat{q}$ like DM, but it improves the reward prediction by $\hat{q}$ using a correction term based on importance sampling. The policy value prediction is computed in this way:
\begin{align*}
    \hat{V}_{DR}(\pi_e; \task, \hat{q}) \coloneqq \frac{1}{n_b} \sum_{t=1}^{n_b} \left( w(x_t, a_t) (r_t - q(x_t, a_t)) + \sum_{a \in \mathcal{A}} \pi_e(a|x_t) \hat{q}(x_t, a) \right) \ \ .
\end{align*}
It has useful statistical properties. It is unbiased, consistent, and in general has a lower variance with respect to IPS.

\paragraph{Self-Normalized Doubly Robust (SNDR)}
The SNDR estimator \citep{dr1, snips1, snips2} is a variant of DR that shrinks the weights using the same normalization factor used in SNIPS. The policy value prediction is computed in this way:
\begin{align*}
    \hat{V}_{SNDR}(\pi_e; \task, \hat{q}) \coloneqq \sum_{t=1}^{n_b} \left( \frac{w(x_t,a_t)}{\sum_{j=1}^n w(x_j,a_j)} (r_t - q(x_t, a_t)) + \frac{1}{n_b}\sum_{a \in \mathcal{A}} \pi_e(a|x_t) \hat{q}(x_t, a) \right)  \ \ .
\end{align*}
As in SNIPS, the normalization factor is used to gain stability, but some bias is introduced.

\paragraph{Subgaussian Importance Sampling Doubly Robust (DR-$\lambda$)}
The DR-$\lambda$ estimator \citep{subgaus} is a variant of DR that shrinks the weights using the same smooth subgaussian by IPS-$\lambda$. The policy value prediction is computed in this way:
\begin{align*}
    \hat{V}_{DR-\lambda}(\pi_e; \task, \hat{q}, \lambda, \gamma) \coloneqq \frac{1}{n_b} \sum_{t=1}^{n_b} \left( w_{\lambda, \gamma}(x_t, a_t) (r_t - q(x_t, a_t)) + \sum_{a \in \mathcal{A}} \pi_e(a|x_t) \hat{q}(x_t, a) \right) \ \ .
\end{align*}
The $\lambda,\ \gamma$ hyperparameters balance the bias-variance trade-off of the estimator.

\paragraph{Doubly Robust with Optimistic Shrinkage (DRos)}
The DRos estimator \citep{dros} is a variant of DR that reduces variance with an importance weight derived by minimizing the sharp bound of estimator MSE. The policy value prediction is computed in this way:
\begin{align*}
    \hat{V}_{DRos}(\pi_e; \task, \hat{q}, \lambda) \coloneqq \frac{1}{n_b} \sum_{t=1}^{n_b} \left( \frac{\lambda w(x_t,a_t)}{w(x_t,a_t)^2 + \lambda} (r_t - q(x_t, a_t)) + \sum_{a \in \mathcal{A}} \pi_e(a|x_t) \hat{q}(x_t, a) \right), 
\end{align*}
where the $\lambda \geq 0$ hyperparameter balances the bias-variance trade-off of the estimator.

\paragraph{Switch}
The Switch estimator \citep{switch} is a variant of DR that switches between DM and DR, using DM when the importance ratio is below a threshold specified by the value of $\lambda$, and DR when $w(x_t,a_t)$ is above the threshold.
The policy value prediction is computed in this way:
\begin{align*}
    & \hat{V}_{\mathrm{switch}}(\pi_e; \task, \hat{q}, \lambda)  \coloneqq 
    \\ & \frac{1}{n_b} \sum_{t=1}^{n_b} \left( w(x_t,a_t) \mathbb{I} \left\{ w(x_t,a_t) \leq \lambda \right\} \cdot (r_t - q(x_t, a_t)) + \sum_{a \in \mathcal{A}} \pi_e(a|x_t) \hat{q}(x_t, a) \right),
\end{align*}
where the $\lambda \geq 0$ hyperparameter balances the bias-variance trade-off of the estimator.

\section{SYNTHETIC DATASET}
\label{app:bb-dataset}
\subsection{Characteristics}
As stated in Section \ref{sec:bb-method}, \bbAutoOPE method is based on a zero-shot approach framed as a supervised learning problem. In particular, the goal of this supervised learning problem is to predict, given an estimator $\hat{V}_m$ and an OPE task $\task^{(j)}$ as input, the $\text{MSE}$ of $\hat{V}_m$ when applied to $\task^{(j)}$. 

Hence, to train a Random Forest regressor to predict this MSE, we need a finite set of possible estimators $\mathcal{V}$ such that $\hat{V}_m \in \mathcal{V}$ defined a-priori, and a large set of OPE tasks $\mathcal{M}$.
The set $\mathcal{M}$ is a collection of synthetically generated OPE tasks, obtained from a data generator function $G$, which we describe in further detail in Section
\ref{app:synth-gen}.

The training dataset for the Random Forest is obtained extracting suitable input features from the estimator $\hat{V}_m$ and an OPE task $\task^{(j)}$, that the Random Forest can work with. This is done using a feature extraction function $g$ (described in Appendix \ref{app:feat-eng}.).
Then, the target is the ground-truth $\text{MSE}$ of $\hat{V}_m$ when applied to $\task^{(j)}$. The training dataset can be expressed as: $\{ \mathrm{input:}\ g(\mathcal{V} \times \mathcal{M}),\ \mathrm{target:}\ \text{MSE}(\mathcal{V} \times \mathcal{M})\}$

The meta-dataset $\mathcal{M}$ contains $250,000$ OPE tasks, and for each task $\task^{(j)}$, $n_{gen} = 10$ realizations $\{\task^{(j)}(s)\}_s^{n_{gen}}$ are produced, for a total of $2,500,000$ OPE tasks. Each task is associated to all the candidate estimators in $\mathcal{V}$ that are $21$, for a total of $52.5$ million samples. Note that we removed the few trivial OPE tasks which had all rewards equal to 0 or 1. 

\paragraph{OPE Estimators}
\label{par:synt-ope-est}
The set $\mathcal{V}$ of candidate estimators includes several model-free, model-based and hybrid estimators, as done in \citep{saito23}. Estimators hyperparameters, when present, were selected using the SLOPE method \citep{slope}. All the estimators used are listed in Table \ref{tab:estimators}, and a brief explanation of each of them is reported in appendix \ref{app:ope-est}.

Each of the six estimators with a reward model is employed in conjunction with three reward predictors: \textit{Random Forest Classifier} and \textit{Logistic Regression} from the \textit{Scikit-Learn}\footnote{Scikit-Learn available with 3-Clause BSD license: \url{https://github.com/scikit-learn/scikit-learn}} library \citep{scikit-learn}, and \textit{LGBM Classifier} from the Python package of \textit{LightGBM}\footnote{LightGBM available with MIT license: \url{https://github.com/microsoft/LightGBM/tree/master}} \citep{lgbm}. The reward models were fitted using a 3-fold cross-fitting procedure to reduce possible overfitting \citep{narita}.

\begin{table}
    \centering
    \small
    \begin{tabular}{c p{5cm}}
        \toprule
        \bfseries Type &\bfseries OPE Estimator \\
        \midrule
        Model-Based & Direct Method (DM) \\
        \hline
        & Inverse Propensity Score (IPS) \\
        Model-Free  & Self-Normalized IPS (SNIPS) \\
        & Sub-Gaussian IPS (IPS-$\lambda$) \\
        \hline
         & Doubly Robust (DR) \\ 
        & Self-Normalized DR (SNDR) \\ 
        Hybrid & Sub-Gaussian DR (DR-$\lambda$) \\ 
        & DR with a optimistic shrinkage (DRos)  \\
        & Switch DR (SwitchDR) \\
        \bottomrule 
    \end{tabular}
    \caption{List of the OPE estimators used in this paper. See Appendix \ref{app:ope-est} for their description. 
    }
    \label{tab:estimators}
\end{table}

    
    



\paragraph{Target}
\label{par:bb-feedbacks}
The target variable is the MSE between the prediction of an estimator $\hat{V}_m(\pi_e; \task^{(j)}(s))\ \text{with}\ \hat{V}_m \in \mathcal{V}$ and the on-policy estimated ground-truth policy value $V(\pi_e)$.
Indeed, together with each synthetic OPE task $\task^{(j)}(s)$, we also generate a ground-truth dataset, denoted as $\mathcal{D}_{gt}^{(j)} \coloneqq \left\{(x_t, a_t, r_t, \left\{\pi_e(a|x_t)\right\}_{a \in \mathcal{A}})\right\}_t^{n_{gt}}$, that is a large CB dataset collected under the evaluation policy $\pie$, which is used to compute the ground-truth MSE, by averaging the collected rewards (see Appendix \ref{app:mse-gt} for further details).


\subsection{Generation}
\label{app:synth-gen}
\paragraph{Data Generator}
\label{par:bb-data-gen}
Synthetizing the previous section, we can say that it is needed a pair $(\task^{(j)}(s), \mathcal{D}_{gt}^{(j)})$ and one estimator $\hat{V}_m$, to generate one row of the dataset on which the Random Forest will be trained.
We can formalize the generation of the pair $(\{\task^{(j)}(s)\}_s^{n_{gen}}, \mathcal{D}_{gt}^{(j)})$ as the output of a function $G$ which serves as synthetic data generator, so $(\task^{(j)}(s), \mathcal{D}_{gt}^{(j)}) = G(p_1^{(j)},\dots,p_P^{(j)})$, with the parameters $(p_1^{(j)},\dots,p_P^{(j)})$ drawn from $\Phi = (\phi_1(\cdot), \dots, \phi_P(\cdot))$.
For each new OPE task, we sample a value for each parameter $p_i$ from the respective distribution $\phi_i$, and generate the datasets pair. Function $G$ is highly configurable since each parameter has a distribution that can be specified, which allows to represent a wide range of scenarios.
Among the parameters $\left\{p_i\right\}_i^P$ that change the OPE task characteristics we include: 
the evaluation and logging policies, the number of possible actions, the number of rounds in the logging data, the context dimension, the reward distribution.

Below we provide the complete list of these parameters, along with their distributions $\left\{\phi_i\right\}_i^P$, providing insights into the characteristics considered for OPE tasks:
\begin{itemize}
    \item $n_{a}$: a positive integer, representing the number of possible actions available in the considered Contextual Bandit scenario. In our generation process, this parameter is distributed as a random uniform in the range $(2, 20)$;  
    \begin{align*}
        n_{a} \in \mathbb{N},\ n_{a} \sim \mathcal{U}(\left\{2, 3, \dots, 20\right\}) \ \ .
    \end{align*}
        
    \item $n_{b}$: a positive integer number representing the number of observed contexts, \ie the number of contexts in the data $\task$. In our generation process, this parameter is distributed as a random uniform in the range $(100, 8000)$; 
    \begin{align*}
        n_b \in \mathbb{N},\ n_{b} \sim \mathcal{U}(\left\{100, 101, \dots, 8000\right\}) \ \ .
    \end{align*}
    \item \textit{$d_{x}$}: a positive integer number representing the context vector dimension. In our generation process, this parameter is distributed as a random uniform in the range $(1, 10)$; 
    \begin{align*}
        d_{x} \in \mathbb{N},\ d_{x} \sim \mathcal{U}(\left\{1, 2, \dots, 10\right\}) \ \ .
    \end{align*}
    
    \item $q$: it can be either a function provided by the generation module of OBP\footnote{\textit{Open Bandit Pipeline} (OBP) is an open-source library specifically developed to establish a standardized pipeline tailored for research in Off-Policy Evaluation. It is released under the Apache 2.0 License (\url{https://github.com/st-tech/zr-obp})}, or \textit{None}.
    This parameter identifies a function that determines how the expected reward is computed given the pair of context and action. In our generation process, this parameter is distributed as a random uniform over the discrete finite space 
    \begin{align*}
        \mathbb{F}_{r} \coloneqq \{\textit{logistic\_reward\_function},\ \textit{logistic\_polynomial\_reward\_function}, \\ 
        \textit{logistic\_sparse\_reward\_function},\ \textit{None}\}, 
    \end{align*}
    \begin{align*}
        q \sim \mathcal{U}(\mathbb{F}_{r}) \ \ .
    \end{align*}

    Given a context vector $x_t \in \mathbb{R}^{d_x}$ and an action $a_t \in \left\{0, 1, \dots, n_a - 1\right\}$, these functions are used to define the mean reward function $q(x_t, a_t) = \mathbb{E}\left[r_t|x_t,a_t\right]$. 
    
    If $q$ is $\textit{None}$, the expected reward is defined as $\mathbb{E}\left[r_t|x_t,a_t\right] \sim \mathcal{U}(0,1)$, while if $q$ is not $\textit{None}$, the definition is more complex. Firstly we need to introduce the symbol $c_t \in \mathbb{R}^{n_a}$, that is the $a_t$-th canonical basis vector relative to the space $\mathbb{R}^{n_a}$. Then we can define 
    \begin{align} 
    \label{eq:synt-rwd-fun} 
        q(x_t, a_t) \coloneqq \sigma(\tilde{x}_t^T M_{X,A} \ \tilde{a}_t + \theta_x^T \tilde{x}_t + \theta_a^T \tilde{a}_t) \ \ ,
    \end{align}
    with $\sigma(\cdot)$ that represent the \textit{sigmoid} function, $M_{X,A} \in \mathbb{R}^{p_x \times p_a},\ \theta_x \in \mathbb{R}^{p_x}$, and $\theta_a \in \mathbb{R}^{p_a}$, and:
    \begin{itemize}
        \item if $q$ is the $\textit{logistic\_reward\_function}$, the symbols $\tilde{x}_t \in \mathbb{R}^{p_x}$ and $\tilde{a}_t \in \mathbb{R}^{p_a} $, are two feature matrices consisting of all $p_x$ polynomial combinations of the elements of $x_t$, and all $p_a$ polynomial combinations of the elements of $c_t$, with degree less than or equal to $p=1$;
        \item if $q$ is the $\textit{logistic\_polynomial\_reward\_function}$, the symbols $\tilde{x}_t \in \mathbb{R}^{p_x}$ and $\tilde{a}_t \in \mathbb{R}^{p_a}$, are two new feature matrices consisting of all $p_x$ polynomial combinations of the elements of $x_t$, and all $p_a$ polynomial combinations of the elements of $c_t$, with degree less than or equal to $p=3$;
        \item if $q$ is the $\textit{logistic\_sparse\_reward\_function}$, the symbols $\tilde{x}_t \in \mathbb{R}^{p_x}$ and $\tilde{a}_t \in \mathbb{R}^{p_a}$, are two new feature matrices consisting of $p_x$ polynomial combinations of the elements of $x_t$, and $p_a$ polynomial combinations of the elements of $c_t$, with degree less than or equal to $p=1$, randomly selected among all the polynomial combinations. Here the $p_x,\ p_a$ combinations randomly chosen are the 10\% of all the possible polynomial combinations. 
    \end{itemize}

    \item $\beta_{b}$: a pair of real numbers $\beta_b = (\left\{\beta_{b,i}\right\}_{i\in{1,2}})$. Both numbers define one logging policy $\pi_{b,i}$ each. In the Contextual Bandit environment, the first half rounds are drawn under $\pi_{b,1}$, the second half under $\pi_{b,2}$. Each number, together with the outcome of $q$, contributes to the computation of logits values, used to derive the actions probabilities for each logging policy.
    Each parameter $\beta_{b,i}$, referred to as the inverse temperature parameter, uniquely characterizes a policy, signifying the proximity of policy $\pi_{b,i}$ to the optimal policy. Higher $\beta_{b,i}$ values imply closer proximity, $\beta_{b,i} = 0$ means a random policy, and negative values denote a greater divergence from the optimal actions distribution.
    In this general formulation, in our generation process, this parameter is distributed as a 2-dimensional vector, with each component distributed as a random uniform with range $(-10, 10)$.
    \begin{align*}
        \beta_b \in \mathbb{R}^2,\ \beta_b \sim (\mathcal{U}(-10, 10),\ \mathcal{U}(-10, 10)) \ \ .
    \end{align*}
    In our data generation, we decided to generate half of the OPE tasks with the parameter $\beta_b$ as a unique real number. This means that half of the synthetic OPE tasks have only one logging policy, while the other half is collected by two logging policies.

    \item $\beta_{e}$: a real number. Analogously to $\beta_b$, it defines the evaluation policy, and contributes to the computation of logits values, used to derive the counterfactual actions probabilities, together with the outcome of $q$. In our generation process, this parameter is distributed as a random uniform with range $(-10, 10)$.
    \begin{align*}
        \beta_e \in \mathbb{R},\ \beta_e \sim \mathcal{U}(-10, 10) \ \ .
    \end{align*} 
    
    \item $f_{\pi_b}$: it can be either a function provided by the generation module of OBP, or \textit{None}.
    This parameter governs the logging policy of our synthetic dataset. We define the logging policy in terms of $f_{\pi_b}$ as follows: $\pi_b(a_t|x_t) \coloneqq \textit{softmax}_{a_t}(\beta_b \cdot f_{\pi_b}(x_t, a_t))$. In our generation process, this parameter is distributed as a random uniform over the discrete finite space $\mathbb{F}_{\pi}$.
    \begin{align*}
        \mathbb{F}_{\pi} & \coloneqq \left\{\textit{polynomial\_behavior\_policy},\ \textit{linear\_behavior\_policy},\ \textit{None} \right\} \ \ ,
    \end{align*}
    \begin{align*}
        f_{\pi_b} \sim \mathcal{U}(\mathbb{F}_{\pi}) \ \ .
    \end{align*}
    
    If $f_{\pi_b}$ is $\textit{None}$, 
    then, we set   $f_{\pi_b}  \coloneqq q(x_t, a_t) $;
    while if $f_{\pi_b}$ is not $\textit{None}$, we use a more complex logging policy. First, we need to introduce the symbol $c_t \in \mathbb{R}^{n_a}$, that is the $a_t$-th canonical basis vector relative to the space $\mathbb{R}^{n_a}$. Then, we define 
    \begin{align} \label{eq:synth-policy-gen}
    f_{\pi_b}(x_t, a_t)  \coloneqq \tilde{x}_t^T M_{X,A} \ \tilde{a}_t + \theta_a^T \tilde{a}_t \ \ ,
    \end{align}
    with $M_{X,A} \in \mathbb{R}^{p_x \times p_a},\ \theta_x \in \mathbb{R}^{p_x}$, and $\theta_a \in \mathbb{R}^{p_a}$, and:
    \begin{itemize}
        \item if $f_{\pi_b}$ is the $\textit{linear\_behaviour\_policy}$, the symbols $\tilde{x}_t \in \mathbb{R}^{p_x}$ and $\tilde{a}_t \in \mathbb{R}^{p_a}$, are two feature matrices consisting of all $p_x$ polynomial combinations of the elements of $x_t$, and all $p_a$ polynomial combinations of the elements of $c_t$, with degree less than or equal to $p=1$; 
        \item if $f_{\pi_b}$ is the $\textit{polynomial\_behaviour\_policy}$, the symbols $\tilde{x}_t \in \mathbb{R}^{p_x}$ and $\tilde{a}_t \in \mathbb{R}^{p_a}$, are two feature matrices consisting of all $p_x$ polynomial combinations of the elements of $x_t$, and all $p_a$ polynomial combinations of the elements of $c_t$, with degree less than or equal to $p=3$.
    \end{itemize}
    
    \item $f_{\pi_e}$: like \textit{$f_{\pi_b}$}, it can be either a function provided by the generation module of OBP, or \textit{None}. This parameter governs the evaluation policy. In our generation process, this parameter is distributed as a random uniform over the discrete finite space $\mathbb{F}_{\pi}$. The functions in $\mathbb{F}_{\pi}$ are described by Equation \ref{eq:synth-policy-gen}.
    \begin{align*}
        \mathbb{F}_{\pi} & \coloneqq \left\{\textit{polynomial\_behavior\_policy},\ \textit{linear\_behavior\_policy},\ \textit{None} \right\} \ \ ,
    \end{align*}
    \begin{align*}
        f_{\pi_e} \sim \mathcal{U}(\mathbb{F}_{\pi}) \ \ .
    \end{align*}
\end{itemize}

Apart from these randomly sampled parameters, the synthetic data generator $G$ has additional parameters that are fixed and independent from the current generated OPE task. These include:
\begin{itemize}
    \item $n_{gen}$: a positive integer determining the number of realizations of each $\task$. In our generation process, $n_{gen} = 10$. 
    \item $n_{gt}$: a positive integer regulating the number of rounds in $\mathcal{D}_{gt}$. In our generation process, $n_{gt} = 100,000$
\end{itemize}

\paragraph{OPE Task Generation}
\label{subapp:ope-data-gen}
Introduced the parameters used in the generation process, the procedure for obtaining the pair $(\{\task^{(j)}(s)\}_s^{n_{gen}}, \mathcal{D}_{gt}^{(j)})$ can be formally described.

The procedure is outlined by Algorithm \ref{alg:bb-data-gen}, which internally calls the function \textit{Generate} and the function \textit{GenerateOPE}, explained in the following:
\begin{description}
    \item[\textit{Generate}]:
    
\begin{enumerate}[label=(\arabic*)]
    \item $n_{gt}$ vectors $x_t$ are independently sampled from a $d_{x}-$dimensional random vector, distributed as a normal with zero mean and identity covariance matrix. 
    \item Then for each context, we compute the expected reward for all available $n_a$ actions. These expected rewards are determined using the synthetic reward function $q$: it is a deterministic function that has the role to generate synthetically $\mathbb{E}_{p(r_t|x_t,a_t)}\left[r_t\right]$.
    \item The evaluation policy $\pi_e(a_t|x_t)$ is defined by this formula $\pi_e(a_t|x_t) = \textit{softmax}_{a_t}(\beta_e \cdot f_{\pi_e}(x_t, a_t))$, and it can be computed for each context $x_t$ and each admissible action \(a_t\).
    \item Then actions $a_t$ are sampled for each context following policy $\pi_e$, and the relative rewards $r_t$ for each pair $(x_t, a_t)$ are sampled from a Bernoulli distribution with a mean of $q(x_t, a_t)$.
    \item The data produced are sufficient to formally define a logging dataset to compute the on-policy value estimate as rewards average.
\end{enumerate}


    \item[\textit{GenerateOPE}]:
    
\begin{enumerate}[label=(\arabic*)]
    \item other $n_{b}$ vectors $x_t$ are sampled from the same $d_{x}$-dimensional normal vector with zero mean and identity covariance matrix.
    \item The logging policy $\pi_b$ is computed for each context $x_t$ and all the available actions. 
    \item An action $a_t$ is then selected for each context $x_t$, following the policy $\pi_b$. 
    \item The correspondent reward $r_t$ is produced and collected, sampling from the same Bernoulli distribution we introduced before.
    \item The data produced are sufficient to formally define $\task^{(j)}(s)$ as in Equation \ref{eq::D-ope}.
\end{enumerate}

\end{description}

\begin{algorithm}
\small
\caption{OPE Data Generator $G$} \label{alg:bb-data-gen}
\begin{algorithmic}[1]
\REQUIRE 
number of realizations $n_{gen}$ of each OPE dataset, a big number of rounds $n_{gt}$ to sample under $\pi_e$ for the evaluation dataset, space of generation parameters $\Phi$
\ENSURE Collection of OPE datasets and evaluation datasets $\{\mathcal{T}^{(j)}_{ope}(s)\}_{s}^{n_{gen}},\ \mathcal{D}^{(j)}_{gt}\}_{j=1}^{n_{data}}$
    \STATE $(n_a, n_b, d_x, q, \beta_b, \beta_e, f_{\pi_b}, f_{\pi_e}) \sim \Phi$
    \STATE $\mathcal{A} \leftarrow \left\{0,\dots,n_a - 1\right\}$
    \STATE $\mathcal{D}^{(j)}_{gt} \leftarrow \mathrm{Generate} (\mathcal{A}, d_x, \beta_e, f_{\pi_e}, q, n_{gt})$
    \FOR{$s = 1,\ldots,n_{gen}$} 
        \STATE $\mathcal{T}^{(j)}_{ope}(s) \leftarrow \mathrm{GenerateOPE} (\mathcal{A}, d_x,\beta_b,\beta_e, f_{\pi_b}, f_{\pi_e}, q, n_b, s)$ 
    \ENDFOR 
\end{algorithmic}
\end{algorithm}

\section{SUPERVISED MODEL TRAINING}
\label{app:bb-model}

\paragraph{Supervised Model} As a supervised model, we select the \textit{Random Forest regressor} provided by the \textit{scikit-learn} package\footnote{\url{https://scikit-learn.org/stable/modules/generated/sklearn.ensemble.RandomForestRegressor.html}}. The learning pipeline comprehend also some preprocessing steps and transformations:
\begin{itemize}
    \item Categorical features are one-hot encoded.
    \item Numerical features are clipped to a maximum value of $10^{10}$.
    \item The function $log(1+x)$ is applied to positive features with a skewness higher than 1 (and to the target variable).
    \item Each numerical feature is scaled to the interval $(0,1)$ if the feature is positive and to $(-1,1)$ if can assume negative values. The scaling is applied also the target variable.
\end{itemize}

\paragraph{Data Split}
\label{par:bb-prep}
In order to train and evaluate the \bbAutoOPE supervised model, we split the meta-dataset into three parts, 60\% for training, 20\% for validation, and 20\% for testing.

\paragraph{Hyperparameter Optimization}
\label{par-bb-hyperparams}
\bbAutoOPE uses a Random Forest as a supervised model. In order to select the best hyperparameters we rely on a Bayesian Search optimizing the \textit{Regret} between the ground-truth MSE of the OPE estimator selected by \bbAutoOPE, $\hat{V}_{\hat{m}}$, and the best estimator according to the MSE on the ground-truth dataset.
We define the Regret as: 
\begin{align}
    \text{Regret} \coloneqq \text{MSE}(\hat{V}_{\hat{m}}(\pi_e)) - \text{MSE}(\hat{V}_{m^*}(\pi_e)) \ ,
\end{align}

This means that at each step of the Bayesian Search, a Random Forest regression model (with fixed hyperparameters) is fitted on the training set, and then is evaluated on the validation set in terms of Regret. 
This Regret is the objective function that the Bayesian Search tries to optimize. 
Finally, the Random Forest is trained on the union of the training and validation set with the best hyperparameters found by the Bayesian Search, and is evaluated on the test set to asses the performance.

The Bayesian Search is implemented on top of the \textit{Scikit-Optimize}\footnote{Scikit-Optimize available with 3-Clause BSD license: \url{https://github.com/scikit-optimize/scikit-optimize/tree/master} } library, and involves Bayesian optimization using Gaussian Processes to approximate the objective function, making the assumption that function points are distributed as a multivariate Gaussian. For our model, 50 Bayesian Search iterations are performed.

\begin{table*}
    \small
    \centering
    \begin{tabular}{l l l l}
        \toprule
        \bfseries Type & \bfseries Hyperparameter & \bfseries Range & \bfseries Prior \\
        \midrule
        Integer & n\_estimator & (50,\ 500) & - \\
        & max\_depth & (1, 100) & uniform \\
        & min\_samples\_split & (2, 50) & uniform \\
        & min\_samples\_leaf & (1, 50) & uniform \\
        \hline
        Real & max\_samples & (0.01, 1.00) & uniform \\
        & max\_features & (0.10, 1.00) & uniform \\
        \hline
        Categorical & criterion & ["squared\_error"] & - \\
        & oob\_score & [True] & - \\
        \bottomrule 
    \end{tabular}
    \caption{Hyperparameters Search Space for \bbAutoOPE Random Forest model Hyperparameters Tuning via Bayesian Search}
    \label{tab:hyperparams}
\end{table*}

\section{FEATURE DESIGN}
\label{app:feat-eng}
This Appendix mathematically describes the input features computed by the feature extractor $g$. They are:
\begin{itemize}
    \item Policy-Independent features: 
    \begin{itemize}
        \item Number of logging rounds, \ie $n_b$.
        \item Number of different possible actions, \ie $n_a$.
        \item Number of deficient actions in logging data, \ie $n_{def} \coloneqq \mid \left\{a \in \mathcal{A} : a \notin \left\{a_t \right\}_t^{n_b}\right\} \mid $.
        \item Context vector dimension, \ie $d_x$.
        \item Variance of logging selected actions, \ie $\sigma^2(a_t) \coloneqq Var[a_t]$.
        \item Mean of logging rewards, \ie $\bar{r}_t \coloneqq \mathbb{E}[r_t]$.
        \item Standard deviation of logging rewards, \ie $\sigma(r_t) = \sqrt{Var[r_t]}$.
        \item Skewness of logging rewards, \ie 
        \begin{align*}
            G_1(r_t) \coloneqq \mathbb{E}\left[\left(\frac{r_t - \mathbb{E}[r_t]}{\sqrt{Var[r_t]}}\right)^3\right].
        \end{align*}
        \item Kurtosis of logging rewards, \ie 
        \begin{align*}
            G_2(r_t) \coloneqq \mathbb{E}\left[\left(\frac{r_t - \mathbb{E}[r_t]}{\sqrt{Var[r_t]}}\right)^4\right].
        \end{align*}
        \item Average variance of context components, \ie 
        \begin{align*}
            \bar{\sigma}^2(x_t) \coloneqq \sum_j^{d_x} Var[x_{t,j}]\ \ .
        \end{align*}
    \end{itemize}
    \item Policy-Dependent features: 
    \begin{itemize}
        \item Max of $\pi_b$ average, over contexts, \ie $\text{max}\ \bar{\pi}_b \coloneqq \text{max}_i\ \frac{1}{n_b} \sum_t^{n_b} \pi_b(a_i|x_t)$. 
        \item Min of $\pi_b$ average, over contexts, \ie $\text{min}\ \bar{\pi}_b \coloneqq \text{min}_i\ \frac{1}{n_b} \sum_t^{n_b} \pi_b(a_i|x_t)$.
        \item Max of $\pi_e$ average, over contexts, \ie $\text{max}\ \bar{\pi}_e \coloneqq \text{max}_i\ \frac{1}{n_b} \sum_t^{n_b} \pi_e(a_i|x_t)$.
        \item Min of $\pi_e$ average, over contexts, \ie $\text{min}\ \bar{\pi}_e \coloneqq \text{min}_i\ \frac{1}{n_b} \sum_t^{n_b} \pi_e(a_i|x_t)$.
        \item Max importance weight, \ie
        \begin{align*}
            \text{max}\ w(\pi_b, \pi_e) \coloneqq \text{max}_t \frac{\pi_b(a_t|x_t)}{\pi_e(a_t|x_t)}\ \ .
        \end{align*}
        \item Average inverse importance weight, \ie
        \begin{align*}
            \bar{w} (\pi_b, \pi_e) \coloneqq \mathbb{E}\left[\frac{\pi_e(a_t|x_t)}{\pi_b(a_t|x_t)}\right]\ \ .
        \end{align*}
        \item Number of importance weight clipped with a threshold $\lambda=10$, \ie
        \begin{align*}
            \#w_{10}(\pi_b, \pi_e) \coloneqq \left\lvert\left\{\forall (x_t,a_t) \in \task,\ w(\pi_b, \pi_e) = \frac{\pi_e(a_t|x_t)}{\pi_b(a_t|x_t)} : w(\pi_b, \pi_e) > 10\right\}\right\rvert\ \ .
        \end{align*}
        \item Total variation distance between $\pi_b, \pi_e$, \ie
        \begin{align*}
            \text{TV}(\pi_b || \pi_e) \coloneqq \frac{1}{n_b} \sum_t^{n_b} \frac{1}{2} \cdot \sum_i^{n_a} \left\lvert \pi_b(a_i|x_t) - \pi_e(a_t|x_t) \right\rvert\ \ .
        \end{align*}
                \item Neyman chi-squared distance between $\pi_b, \pi_e$, which we included because it is equivalent (up to a constant) to the exponentiated \renyi (shown to be an important data characteristic by \citet{metelli-pois}):  
        \begin{align*}
            \text{Neyman}(\pi_b || \pi_e) = \exp(d_r(\pi_e || \pi_b)) - 1\ \ .
        \end{align*}
        We compute it as:
        \begin{align*}
            \text{Neyman}(\pi_b || \pi_e) \coloneqq \frac{1}{n_b} \sum_t^{n_b} \sum_i^{n_a} \frac{(\pi_b(a_i|x_t) - \pi_e(a_i|x_t))^2}{\pi_b(a_i|x_t)}\ \ .
        \end{align*}
        \item Pearson chi-squared distance between $\pi_b, \pi_e$, \ie
        \begin{align*}
            \text{P}(\pi_b || \pi_e) \coloneqq \frac{1}{n_b} \sum_t^{n_b} \sum_i^{n_a} \frac{(\pi_b(a_i|x_t) - \pi_e(a_i|x_t))^2}{\pi_e(a_i|x_t)}\ \ .
        \end{align*} 
        \item Inner product distance between $\pi_b, \pi_e$, \ie
        \begin{align*}
            \text{Inner}(\pi_b || \pi_e) \coloneqq \frac{1}{n_b} \sum_t^{n_b} \sum_i^{n_a} \pi_b(a_i|x_t) \cdot \pi_e(a_i|x_t)\ \ .
        \end{align*}
        \item Chebychev distance between $\pi_b, \pi_e$, \ie
        \begin{align*}
            \text{Chebyshev}(\pi_b || \pi_e) \coloneqq \frac{1}{n_b} \sum_t^{n_b} \text{max}_i \left\lvert \pi_b(a_i|x_t) - \pi_e(a_i|x_t) \right\rvert\ \ .
        \end{align*}
        \item Divergence between $\pi_b, \pi_e$, \ie
        \begin{align*}
            \text{Divergence}(\pi_b || \pi_e) \coloneqq \frac{1}{n_b} \sum_t^{n_b} 2 \cdot \sum_i^{n_a} \frac{(\pi_b(a_i|x_t) - \pi_e(a_i|x_t))^2}{(\pi_b(a_i|x_t) + \pi_e(a_i|x_t))^2}\ \ .
        \end{align*}
        \item Canberra distance metric between $\pi_b, \pi_e$, \ie
        \begin{align*}
            \text{Canberra}(\pi_b || \pi_e) \coloneqq \frac{1}{n_b} \sum_t^{n_b} \sum_i^{n_a} \frac{\left\lvert\pi_b(a_i|x_t) - \pi_e(a_i|x_t)\right\rvert}{\pi_b(a_i|x_t) + \pi_e(a_i|x_t)}\ \ .
        \end{align*}
        \item K-divergence between $\pi_b, \pi_e$, \ie 
        \begin{align*}
            \text{K}(\pi_b || \pi_e) \coloneqq \frac{1}{n_b} \sum_t^{n_b} \sum_i^{n_a} \pi_b(a_i|x_t) \cdot \log\left(\frac{2 \cdot \pi_b(a_i|x_t)}{\pi_b(a_i|x_t) + \pi_e(a_i|x_t)}\right)\ \ .
        \end{align*}
        \item K-divergence between $\pi_e, \pi_b$, note that the policies are swapped, \ie 
        \begin{align*}
            \text{K}(\pi_e || \pi_b) \coloneqq \frac{1}{n_b} \sum_t^{n_b} \sum_i^{n_a} \pi_e(a_i|x_t) \cdot \log\left(\frac{2 \cdot \pi_e(a_i|x_t)}{\pi_e(a_i|x_t) + \pi_b(a_i|x_t)}\right)\ \ .
        \end{align*}
        \item Jensen-Shannon distance between $\pi_b, \pi_e$, \ie
        \begin{equation*}
            \begin{split}
                \text{JS}(\pi_b || \pi_e) \coloneqq \frac{1}{n_b} \sum_t^{n_b} \frac{1}{2} \cdot \left[\sum_i^{n_a} \pi_b(a_i|x_t) \cdot \log\left(\frac{2 \cdot \pi_b(a_i|x_t)}{\pi_b(a_i|x_t) + \pi_e(a_i|x_t)}\right) \right. \\ + \left. \sum_i^{n_a} \pi_e(a_i|x_t) \cdot \log\left(\frac{2 \cdot \pi_e(a_i|x_t)}{\pi_b(a_i|x_t) + \pi_e(a_i|x_t)}\right)\right]\ \ .
            \end{split}
        \end{equation*}
        \item Kullback-Leibler divergence between $\pi_b, \pi_e$, \ie 
        \begin{align*}
            \text{KL}(\pi_b || \pi_e) \coloneqq \frac{1}{n_b} \sum_t^{n_b} \sum_i^{n_a} \pi_b(a_i|x_t) \cdot \log\left(\frac{\pi_b(a_i|x_t)}{\pi_e(a_i|x_t)}\right)\ \ .
        \end{align*}
        \item Kullback-Leibler divergence between $\pi_e, \pi_b$, note that the policies are swapped, \ie
        \begin{align*}
            \text{KL}(\pi_e || \pi_b) \coloneqq \frac{1}{n_b} \sum_t^{n_b} \sum_i^{n_a} \pi_e(a_i|x_t) \cdot \log\left(\frac{\pi_e(a_i|x_t)}{\pi_b(a_i|x_t)}\right)\ \ .
        \end{align*}
        \item Kumar-Johnson distance between $\pi_b, \pi_e$, \ie
        \begin{align*}
            \text{KJ}(\pi_b || \pi_e) \coloneqq \frac{1}{n_b} \sum_t^{n_b} \sum_i^{n_a} \frac{\left((\pi_b(a_i|x_t)^2 - \pi_e(a_i|x_t)^2 \right)^2}{2 \cdot \sqrt{(\pi_b(a_i|x_t) \cdot \pi_e(a_i|x_t))^3}}\ \ .
        \end{align*}
        \item Additive symmetric chi-squared distance between $\pi_b, \pi_e$, \ie
        \begin{align*}
            \text{Additive}(\pi_b || \pi_e) \coloneqq \frac{1}{n_b} \sum_t^{n_b} \sum_i^{n_a} \frac{(\pi_b(a_i|x_t) - \pi_e(a_i|x_t))^2 \cdot (\pi_b(a_i|x_t) + \pi_e(a_i|x_t))}{\pi_b(a_i|x_t) \cdot \pi_e(a_i|x_t)}\ \ .
        \end{align*}
        \item Euclidian distance between $\pi_b, \pi_e$, \ie
        \begin{align*}
            \text{Euclidian}(\pi_b || \pi_e) \coloneqq \frac{1}{n_b} \sum_t^{n_b} \sqrt{\sum_i^{n_a} \left\lvert \pi_b(a_i|x_t) - \pi_e(a_i|x_t)\right\rvert^2}\ \ .
        \end{align*}
        \item Kulczynski similarity between $\pi_b, \pi_e$, \ie
        \begin{align*}
            \text{Kulczynski}(\pi_b || \pi_e) \coloneqq \frac{1}{n_b} \sum_t^{n_b} \frac{\sum_i^{n_a} \left\lvert \pi_b(a_i|x_t) - \pi_e(a_i|x_t)\right\rvert}{\sum_i^{n_a}\text{min}(\pi_b(a_i|x_t),\ \pi_e(a_i|x_t))}\ \ .
        \end{align*}
        \item City-block distance between $\pi_b, \pi_e$, \ie
        \begin{align*}
            \text{CB}(\pi_b || \pi_e) \coloneqq \frac{1}{n_b} \sum_t^{n_b} \sum_i^{n_a} \left\lvert \pi_b(a_i|x_t) - \pi_e(a_i|x_t)\right\rvert\ \ .
        \end{align*}
    \end{itemize}
    \item OPE estimators features: 
    \begin{itemize}
        \item Whether the estimator is a Self-Normalized estimator or not,
        \item Whether the estimator uses Importance Sampling or not,
        \item Whether the estimator uses a Machine Learning reward model or not,
        \item Whether the estimator is a Sub-Gaussian estimator or not,
        \item Whether the estimator is a Shrinked estimator or not,
        \item Whether the estimator is a Switch estimator or not,
        \item Whether the estimator uses a Random Forest Classifier as reward model or not,
        \item Whether the estimator uses a LGBM Classifier as reward model or not,
        \item Whether the estimator uses a Logistic Regression as reward model or not.
    \end{itemize}
\end{itemize}

\begin{figure*}
    \centering
    \subfloat{\includegraphics[scale=0.35]{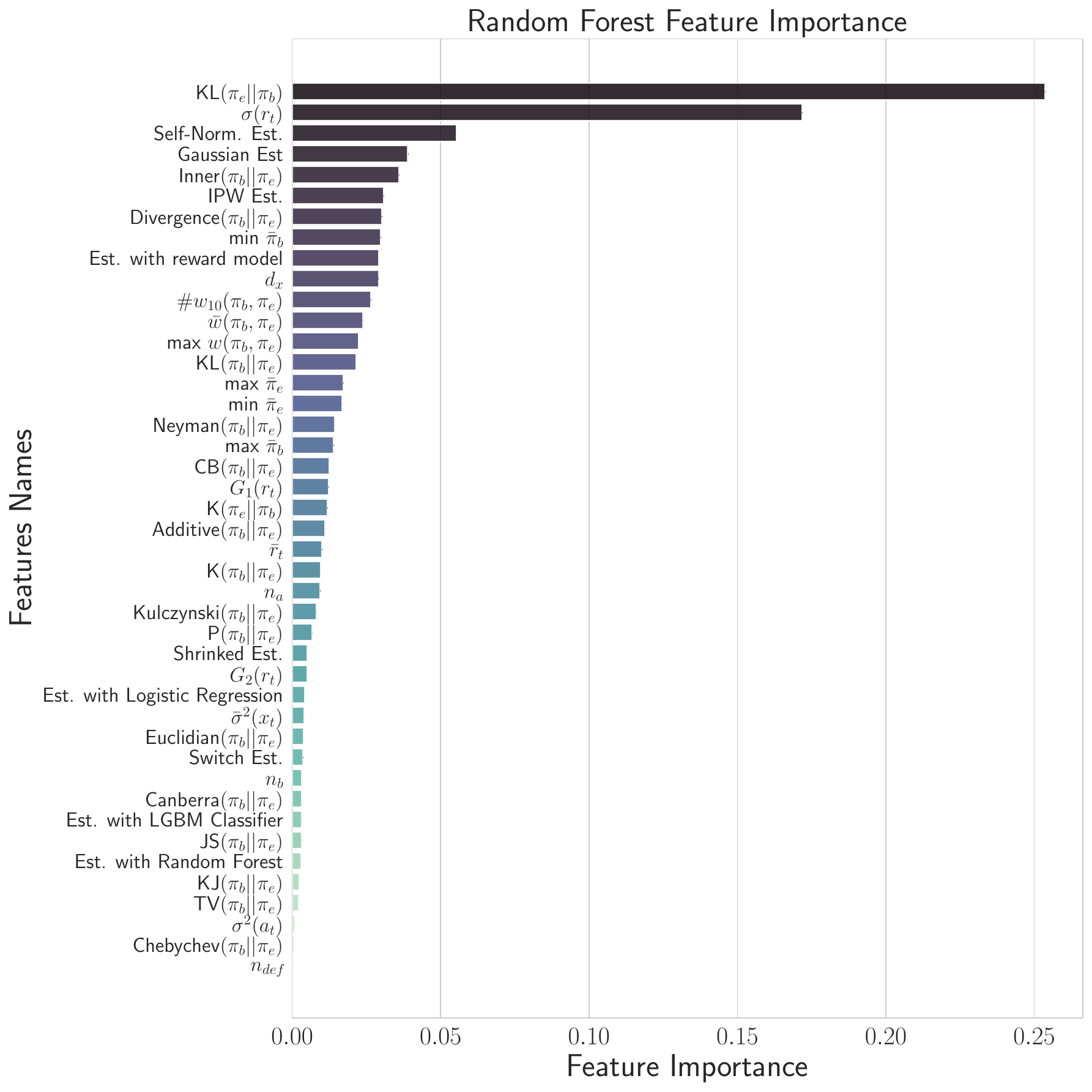}}
    \caption{Plot of the Feature Importance measured with Mean Decrease in Impurity (MDI). The most important features are at the top. 
    }
    \label{fig:full-feature-importance}
\end{figure*}

\paragraph{Feature Importance}
In Figure \ref{fig:full-feature-importance} is reported the complete Feature Importance plot, measured with Mean Decrease in Impurity (MDI).
The 10 policy-independent features appear to be of limited importance and correspond to 5 of the 15 least important features. Only 2 policy-independent features belong to the first 15 most important features.
Policy-dependent features, which are 24, are generally useful for the model, with 9 among the 15 most important features and 6 out of 24 among the 15 least important. 
The KL divergence is the most important feature, which confirms what previous studies in theory have found \citep{chatterjee2018sample}.
The 9 OPE estimator features are similarly spread, with 4 among the most important features and 4 among the least important ones. 

The insights gained from this feature importance analysis is another contribution. In particular, it can shed light on future theoretical findings, lead by analyzing the features that had a significant impact on the model's performance.

\section{ADDITIONAL EXPERIMENTS}
\label{app:additional-exp}
\subsection{Additional Experiments on CIFAR-10}
\label{app:cifar10}
The CIFAR-10 dataset \citep{cifar} is a widely-used benchmark for image classification tasks, with mutually exclusive labeling.

\paragraph{CB Data}
\label{par:cifar10-data}
To adapt the CIFAR-10 dataset for CB we perform the standard supervised to bandit conversion \citep{beygelzimer2009offset, joachims2018deep}, reinterpreting the input features (\ie images pixels scaled in $(0, 1)$) as context features and the class labels as actions.
Furthermore, we partition the dataset into two subsets, the first one is used as logging data $\mathcal{D}_b$ for the OPE task, while the second is used to generate both logging and evaluation policies. In order to generate the policies, the second data split is used to train two Logistic Regression classifiers. Each classifier defines a deterministic policy $\pi_{det}$ because, given a context, it predicts a specific class label (action) with a probability of 1, while all other labels having a probability of 0. 
To produce the final stochastic policies we
blend the deterministic policy with a uniform random policy $\pi_u(a|x)$ using using parameters $\alpha_b$, $\alpha_e$:
\begin{align*}
     \pi_b(a|x) \coloneqq \alpha_b \cdot \pi_{det, b}(a|x) + (1 - \alpha_b) \cdot \pi_u(a|x) \ \ \ , \\
     \pi_e(a|x) \coloneqq \alpha_e \cdot \pi_{det, e}(a|x) + (1 - \alpha_e) \cdot \pi_u(a|x) \ \ \ ,
\end{align*}
with $0 \leq \alpha_e \leq 1 \ , \ \ 0 \leq \alpha_b \leq 1 $. We set $\alpha_b = 0.2$ for the logging policy and $\alpha_e \in \left\{0, 0.25, 0.5, 0.75, 0.99\right\}$ to create diverse evaluation policies, generating different OPE tasks.


\paragraph{Results}
\begin{figure}
     \centering
     \subfloat{\includegraphics[width=0.4\columnwidth]{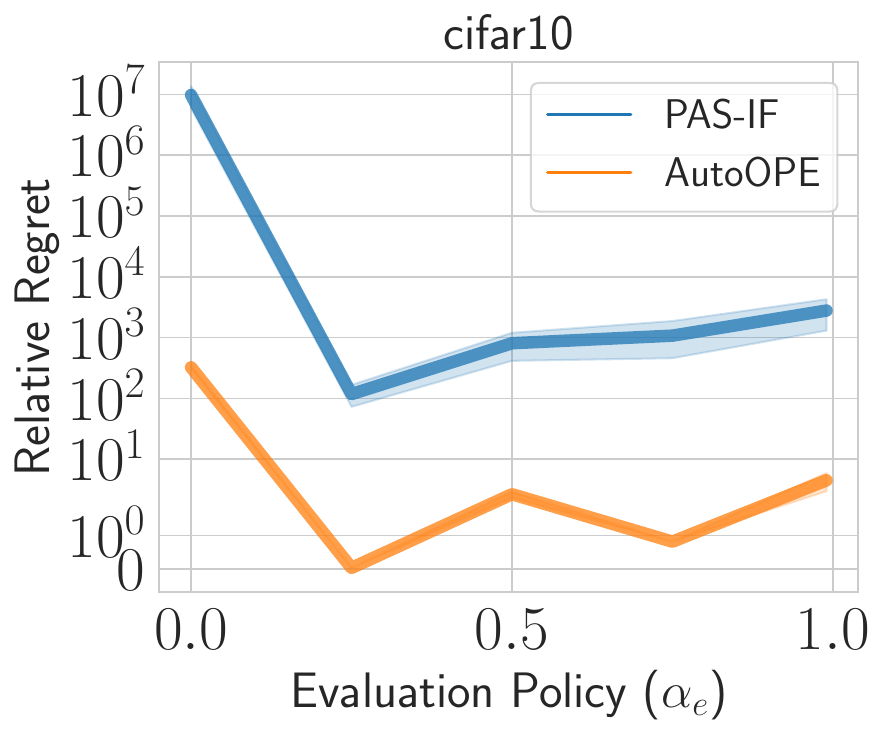}}
     \subfloat{\includegraphics[width=0.4\columnwidth]{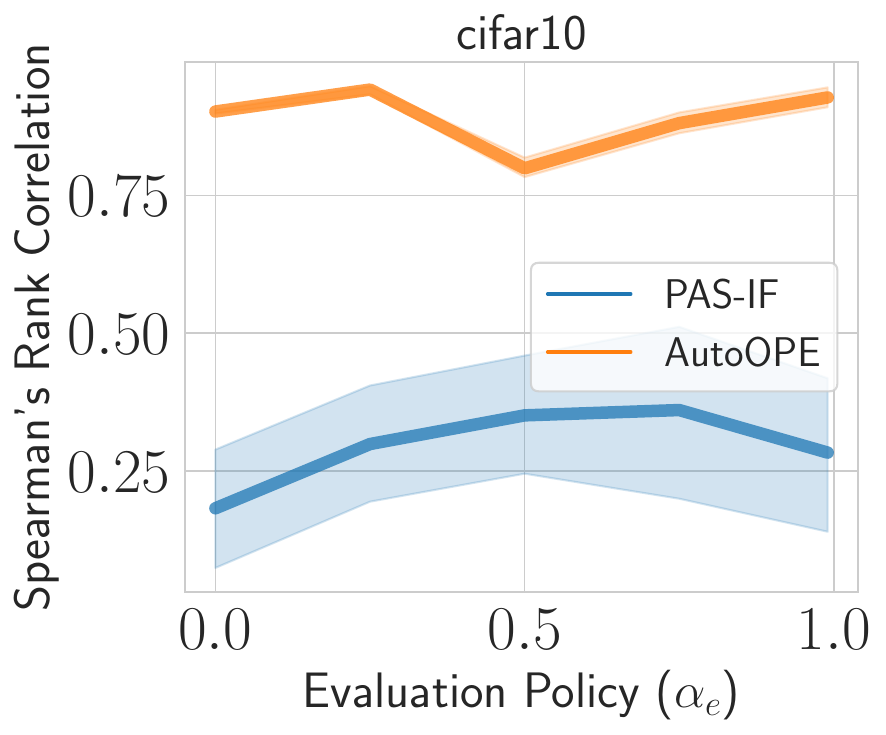}}
     \caption{Relative Regret (lower is better) and Spearman's Rank Correlation Coefficient (higher is better) for Off-Policy Estimator Selection of experiments on CIFAR-10 dataset. Shaded areas correspond to 95\% confidence intervals.}
     \label{fig:cifar10-res}
\end{figure}

In order to be able to compute the variance, we perform 20 bootstrap samples of the logging data 
and we perform the experimental tests on each of them. 

The results are shown in Figure \ref{fig:cifar10-res}. \bbAutoOPE consistently outperforms \pasif, demonstrating lower Relative Regret and better Spearman's Rank Correlation Coefficient. Notably, \bbAutoOPE exhibits a much lower variance in particular for the Relative Regret, while \pasif exhibits very high variance. These results are consistent with the experiments showed in Section \ref{sec:exp} and confirm that \bbAutoOPE is able to generalize to unseen OPE tasks with a different distribution compared to the one used for its training, in a zero-shot scenario.

\subsection{Additional Synthetic Experiments}
\label{app:synt-exp}

\begin{figure}
\centering
\begin{minipage}{0.48\columnwidth}
    \subfloat{\includegraphics[width=0.5\columnwidth]{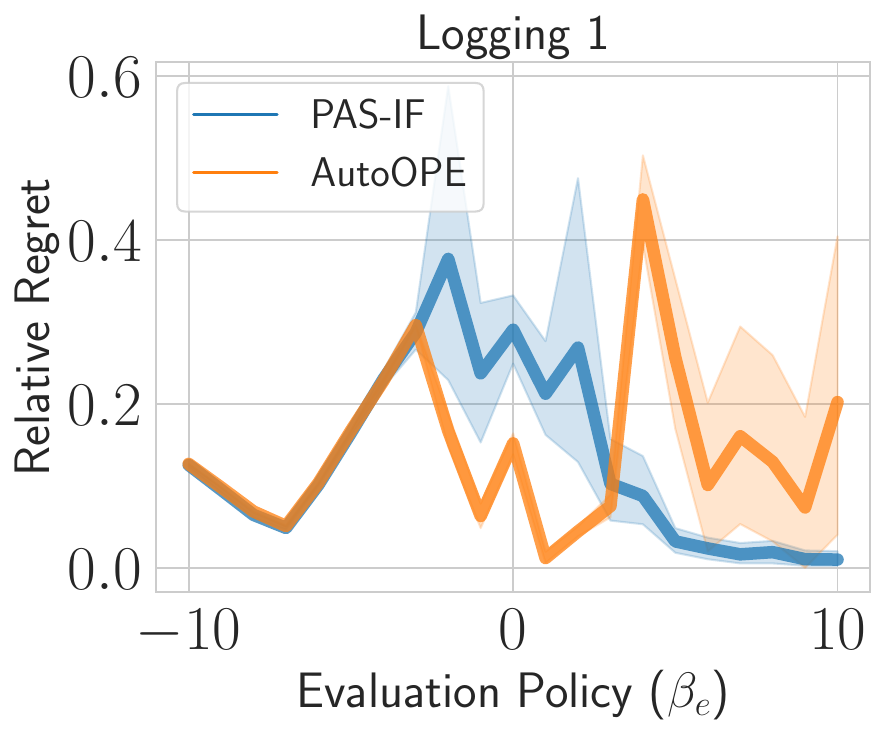}}
    \subfloat{\includegraphics[width=0.5\columnwidth]{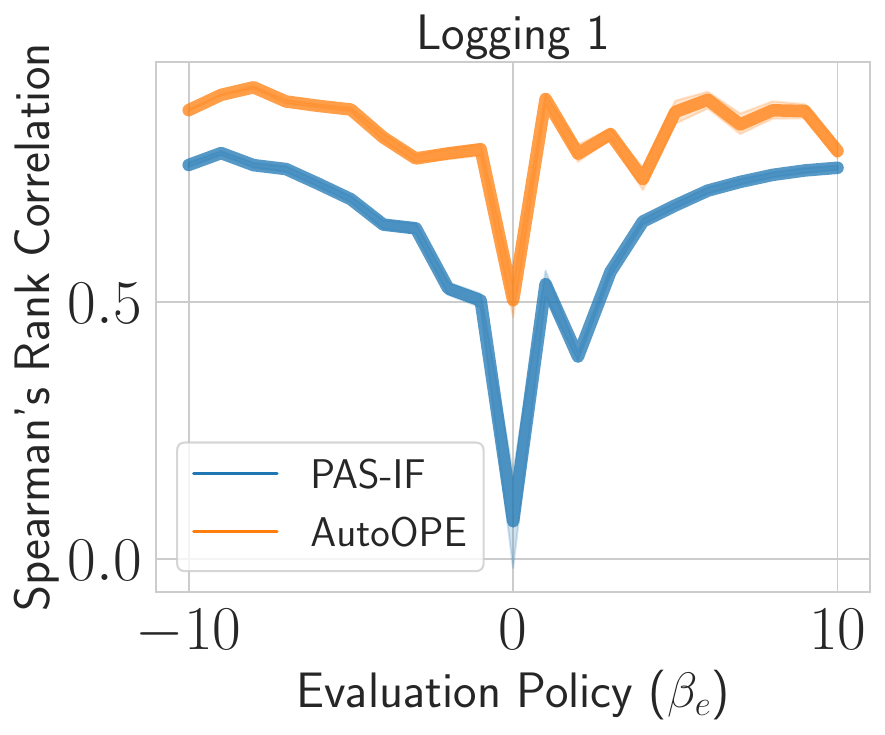}}
    \caption{Relative Regret (lower is better) and Spearman's Rank Correlation Coefficient (higher is better) for Off-Policy Estimator Selection in synthetic experiments with \textbf{Logging 1}. Shaded areas correspond to 95\% confidence intervals.} 
    \label{fig:synt-1}
\end{minipage}
\hfill
\begin{minipage}{0.48\columnwidth}
    \subfloat{\includegraphics[width=0.5\columnwidth]{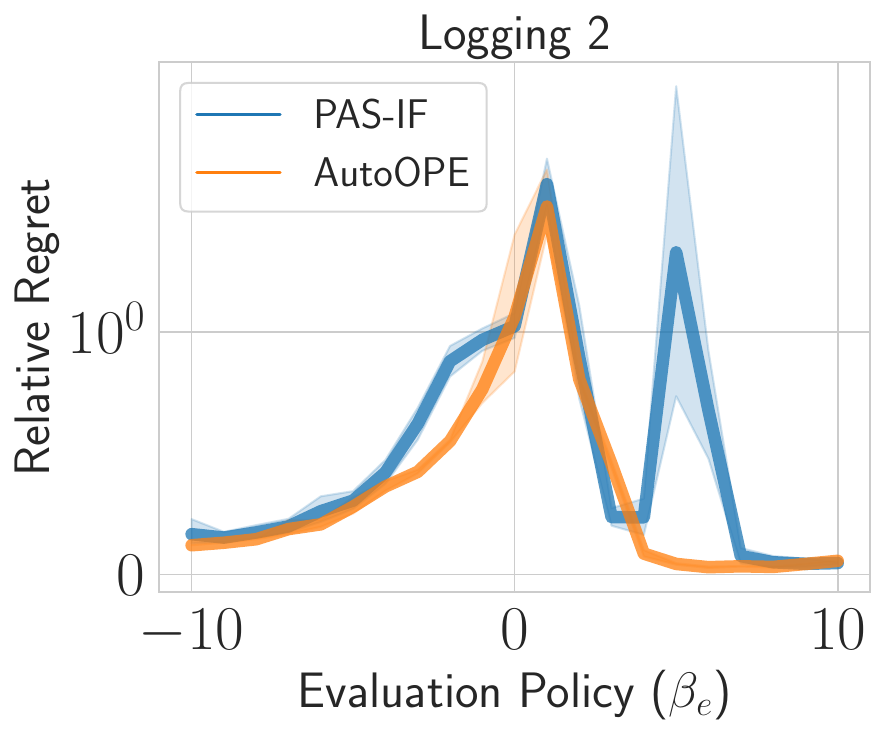}}
    \subfloat{\includegraphics[width=0.5\columnwidth]{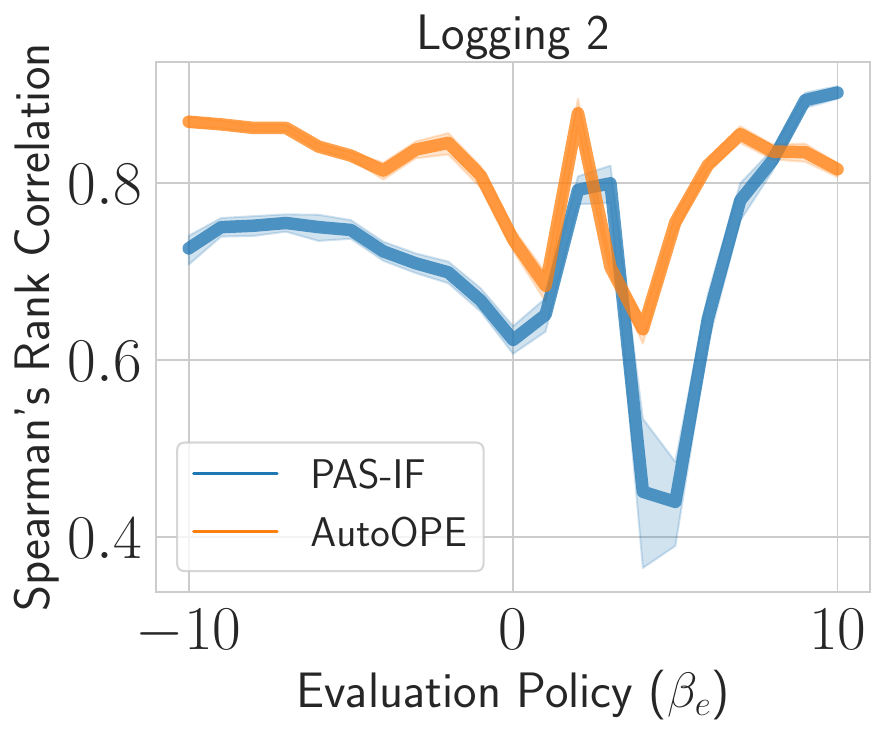}}
    \caption{Relative Regret (lower is better) and Spearman's Rank Correlation Coefficient (higher is better) for Off-Policy Estimator Selection in synthetic experiments with \textbf{Logging 2}. Shaded areas correspond to 95\% confidence intervals.
    }
    \label{fig:synt-2}
\end{minipage}
\end{figure}

In this section we describe all the experiments on synthetic data, following the experimental pipeline of \citep{saito23}. 

\paragraph{CB Data}
\label{par:synth-data}
The synthetic experiments are based on two different logging policies, \textbf{Logging 1} and \textbf{Logging 2}. For each of them 21 OPE tasks with different evaluation policies are defined via a parameter $\beta_e$. 

The synthetic logging data is generated using the following procedure and distributions (same procedure used in \citep{saito23}): 

\begin{itemize}
    \item Context vectors are generated as 10-dimensional vectors, sampled from a multivariate normal distribution with zero mean and identity covariance matrix.
    \item A total of 10 possible actions is considered.
    \item For each context, we compute the expected reward $\mathbb{E}\left[r(x_t, a_t)\right]$, for all available actions. These expected rewards are determined using a synthetic reward function from the `dataset' module of OBP, specifically the `logistic\_reward\_function', described by Equation \ref{eq:synt-rwd-fun}. 
\end{itemize}

The logging policies $\pi_b(a_t|x_t)$ in these experiments are defined as a combination of two policies, each represented as $\pi_i(a_t|x_t) = \textit{softmax}_{a_t}(\beta_i \cdot q(x_t, a_t))$, with $i \in \left\{1, 2\right\}$: the first half of rounds is generated under $\pi_1$, and the second half under $\pi_2$. 
In total, $n_{b,i} = 1,000$ contexts are drawn for each policy $\pi_i$, resulting in a total number of logging rounds $n_{b} = 2000$.

We compute the logging distribution $\pi_b(a_t|x_t)$ of all actions, for each context $x_t$ sampled, and afterwards, we randomly sample an action $a_t$ for each $x_t$, according to this distribution. The actual rewards for these pairs $(x_t, a_t)$ are sampled from a Bernoulli distribution with a mean of $q(x_t, a_t)$. This process yields a series of context-action-reward triplets $\left\{(x_t, a_t, r_t)\right\}_t^{n_b}$. These triplets, together with the action distribution $\pi_b$ computed for all the $x_t$, constitute the logging dataset $\mathcal{D}_b = \left\{(x_t, a_t, r_t, \left\{\pi_b(a|x_t)\right\}_{a \in \mathcal{A}})\right\}_t^{n_b}$.

The evaluation policy $\pi_e$ is described using the same expression as the logging policies, namely $\pi_e(a_t|x_t) = \text{softmax}_{a_t}(\beta_e \cdot q(x_t, a_t))$, and the counterfactual action distribution is computed for each context $x_t$ in $\mathcal{D}_b$. It can be noticed that, with this evaluation policy definition, the $\pi_e$ is uniquely determined by the parameter $\beta_e$.
At the end, the complete OPE dataset generated for a synthetic experiment is described as $\task \coloneqq \left\{\mathcal{D}_b, \left\{\pi_e(a|x_t)\right\}_{a \in \mathcal{A}}\right\}_t^{n_b}$. 

As already said, we divide the synthetic experiments into two main settings: one named \textbf{Logging 1}, where we set $\beta_1 = 2$ and $\beta_2 = -2$, and the other called \textbf{Logging 2}, where $\beta_1 = 3$ and $\beta_2 = 7$, in line with the experimental analysis performed by \citet{saito23}.

By varying $\beta_e$ values in a range from -10 to 10 for both \textbf{Logging 1} and \textbf{2}, we consider 21 different OPE experiments in each setting. 

To provide precise performance assessments, in all the 21 experiments of both \textbf{Logging 1} and \textbf{2} we generate $n_{data} = 100$ realizations of the logging dataset, as done in the procedure outlined in \citep{saito23}. The performance metrics are subsequently averaged over all realizations to yield more robust performance estimates with their confidence intervals.

\paragraph{Results}
\label{par:synt-res}
The results obtained from \textbf{Logging 1} are reported in Figure \ref{fig:synt-1} while those of \textbf{Logging 2} in Figure \ref{fig:synt-2}.  Since these synthetic experiments are the same of \citep{saito23}, we use their optimal values for the other hyperparameters such as the network structure, regularization, learning rate.
In both cases our proposed \bbAutoOPE outperforms \pasif across almost all evaluation policies in terms of Spearman's rank correlation coefficient. 
In terms of Relative Regret, \bbAutoOPE performs similar to \pasif in the first half settings of the Logging 1 experiments. However, for high values of $\beta_e$, \pasif slightly outperform \bbAutoOPE. In the Logging 2 set of experiments, we see that for all the $\beta_e \geq 0$ \bbAutoOPE outperforms \pasif, but \pasif performs better in the settings with $\beta_e < 0$.
The fact that \bbAutoOPE exhibits better Spearman's Rank Correlation indicates that, while both methods predict similar values for the best estimator, \bbAutoOPE is more consistent in ranking all of them.


\section{ABLATION STUDIES}
\label{app:ablation}
In this appendix, we report some ablation studies. Due to the specific settings of these experiments, several new model optimizations need to be executed. Given the limited time and resources, we consider a smaller version of the meta-dataset 
$\mathcal{M}$ to produce results within a reasonable timeframe. For the same reason, we fix the number of trees in the Random Forest to 200, which we find to be a reasonable trade-off between power and manageability.

In the original dataset, each OPE task generated has 10 realizations, with a row of feature values for each realization. In this smaller version of the meta-dataset, however, we consider only one row per OPE task, obtained by averaging the features of the 10 realizations.

These experiments aim to demonstrate that limiting certain characteristics of the meta-dataset's OPE tasks, such as the number of actions or the KL-divergence of the policies, can affect model performance. Importantly, the conclusions obtained from the following experiments remain valid, independently of the meta-dataset size.

\subsection{Scaling Experiment}
\label{sec:scaling_experiment} 
\begin{figure}
     \centering
     \includegraphics[width=.8\columnwidth]{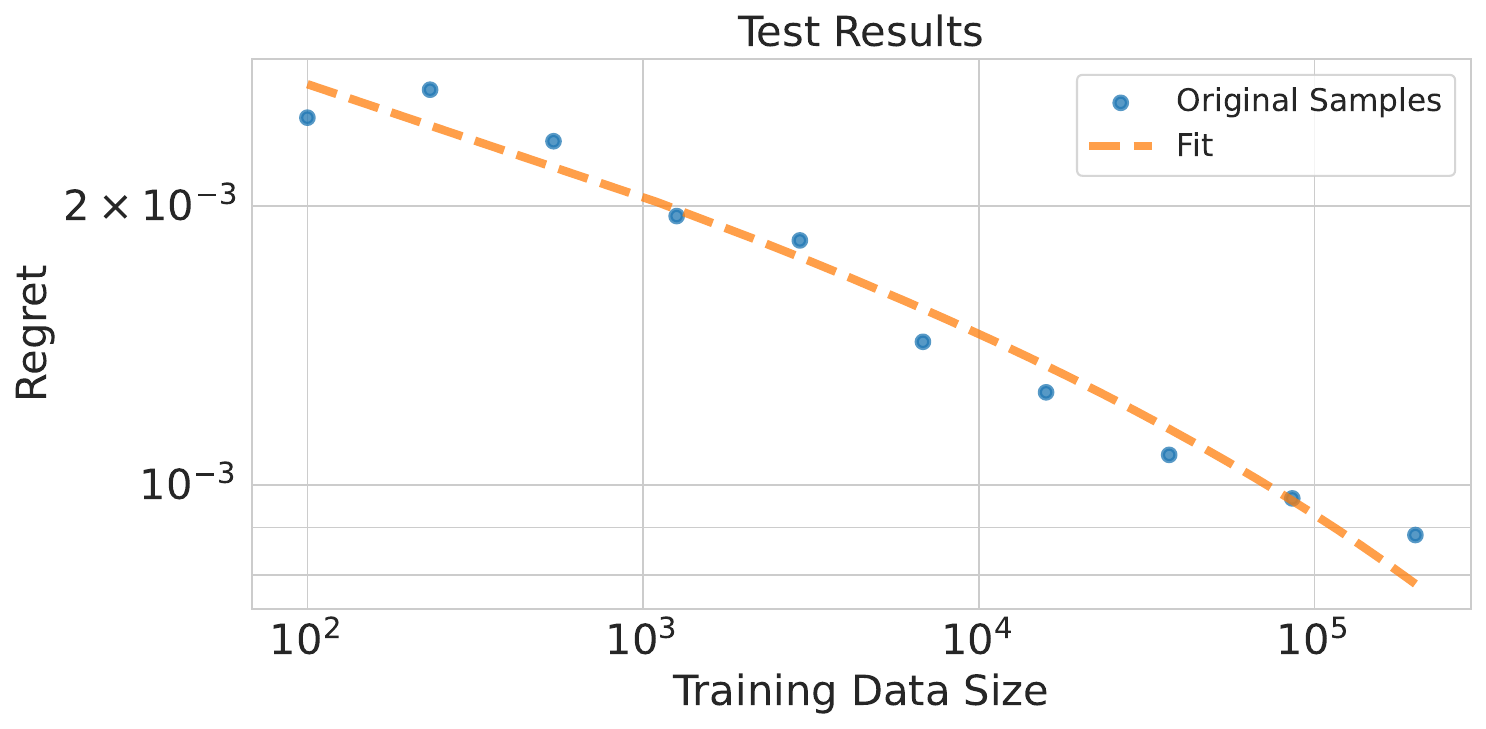}
     \caption{Scaling experiment by varying the meta-dataset size. }
     \label{fig:scaling}
\end{figure}

We performed a scaling experiment of \bbAutoOPE by varying the size of the dataset $\mathcal{M}$. We take 20\% of the OPE tasks in $\mathcal{M}$ as the test set. To investigate the scaling behavior, we train and optimize \bbAutoOPE on different sub-samples of the remaining 80\% of the OPE tasks in $\mathcal{M}$.
The results are reported in Figure \ref{fig:scaling} and show that the loss (\ie the Regret) of \bbAutoOPE follows a scaling behavior, which is particularly popular in the recent Deep Learning literature \citep{kaplan2020scaling, chinchilla}. By increasing the training data, we observe a decrease in the Regret approximately following a power law in the form: $L = A + \frac{B}{D^\alpha}$, where $L$ is the loss measured with the Regret, $D$ is the meta-dataset size, and $A$, $B$, and $\alpha$ are scalar coefficients. 
While we selected the largest meta-dataset for the experimental analysis of this paper, the results in Figure \ref{fig:scaling} show that the dataset size can be adjusted based on the trade-off between computational resources and model performance for the scenario of interest.

\subsection{Additional Ablation Studies}

\paragraph{Features}  To understand the impact of including different feature numbers and types on \bbAutoOPE's performance, we conducted the following ablation study. We group the features in the 3 sets defined in the manuscript:
\begin{itemize}
    \item policies-dependent;
    \item policies-independent;
    \item estimators-dependent.
\end{itemize}
 We train 3 models (one for each group) in the same way outlined in Appendix \ref{app:bb-model}, and we test them on the same unseen test set used for the scaling experiment. We show the results in Table \ref{tab:ablation-features}.

 \begin{table}
     \centering
    $$
    \begin{array}{l|l|l}
    \hline
    \text{Features subset} &  \text{Rel. Regret} (\downarrow) & \text{Spearman} (\uparrow) \\
    \hline
    \text{Estimators-dependent} & 13.58 & 0.23 \\
    \text{Policies-independent} & 5.22 & 0.35 \\
    \text{Policies-dependent} & 5.14 & 0.43 \\
    \text{All} & \mathbf{2.41} & \mathbf{0.48} \\
    \hline
    \end{array}
    $$
     \caption{Ablation study on the features of the model.}
     \label{tab:ablation-features}
 \end{table}
 
 From these results, we can see the impact of each feature group: using only estimator-dependent features gives poor results, policies-dependent features are the best individual group of features, and using all the features combined is beneficial. This applies to both regret and Spearman.

\paragraph{Dataset diversity}
In order to understand the impact of the meta-dataset characteristics on the performance of \bbAutoOPE, we performed the following ablation study.
Here, we simulate different generation methods by subsampling the complete dataset $\mathcal{M}$. For this study, we perform two experiments:
\begin{itemize}
    \item First, we keep only the OPE tasks with 5 actions or less and train only on these. This experiment should highlight the importance of having tasks with a higher number of actions in the meta-dataset. To compare this generation method with the one used in the paper, we train a model on a random subset of $\mathcal{M}$ with the same size. The results obtained are reported in Table \ref{tab:ablation-actions}.
    \item In the second experiment, we only keep the tasks with a KL-divergence between the policies $\leq 0.1$. This experiment should highlight the importance of having tasks where the policies are different in the meta-dataset. To compare this generation method with the one used in the paper, we train a model on a random subset of $\mathcal{M}$ with the same size. The results obtained are reported in Table \ref{tab:ablation-policies}.
\end{itemize}

\begin{table}
    \centering
    $$
    \begin{array}{l|l|l}
    \hline
    \text{Generation Methods} & \text{Rel. Regret}(\downarrow) & \text{Spearman} (\uparrow) \\
    \hline
    |A|\leq 5 & 7.39 & 0.33 \\
    \text{All Tasks} & \mathbf{3.66} & \mathbf{0.45} \\
    \hline
    \end{array}
    $$
    \caption{Ablation study on the number of actions in the dataset.}
    \label{tab:ablation-actions}
\end{table}

\begin{table}
    \centering
    $$
    \begin{array}{l|l|l}
    \hline
    \text{Generation Methods} & \text{Rel. Regret} (\downarrow) & \text{Spearman} (\uparrow)\\
    \hline
    \text{KL}\leq 0.1 & 117.48 & 0.15 \\
    \text{All Tasks} & \mathbf{16.73} & \mathbf{0.43} \\
    \hline
    \end{array}
    $$
    \caption{Ablation study on the type of policies in the dataset.}
    \label{tab:ablation-policies}
\end{table}

From these findings, we can see how a more diverse generation method is beneficial and improves performance compared to a generation method that generates tasks with a low number of actions or with similar logging and evaluation policies.

\section{INFRASTRUCTURE AND COMPUTATIONAL TIME COMPARISON}
\label{app:infrastructure}
For the experiments conducted in this paper, we employed different instances from AWS EC2, to match different needs of resources (GPUs, CPUs) according to the different experiments. The Operating System Ubuntu 20.04 was selected for all the tests.

We used the \texttt{c6i.32xlarge} instance with 128 cores and 256 GB RAM for the synthetic dataset generation explained in Appendix \ref{par:bb-data-gen}. This instance allows to terminate the whole generation process in about 35 hours. 

For the \bbAutoOPE model optimization, training and evaluation instead, additional RAM is needed, since some optimization iterations dive into the training of many and very deep decision trees. The instance used in this case is the \texttt{m6i.32xlarge} instance, with 128 cores and 512 GB RAM. The whole process last approximately 5 days.

\begin{figure}
\begin{minipage}{\columnwidth}
    \includegraphics[width=\columnwidth]{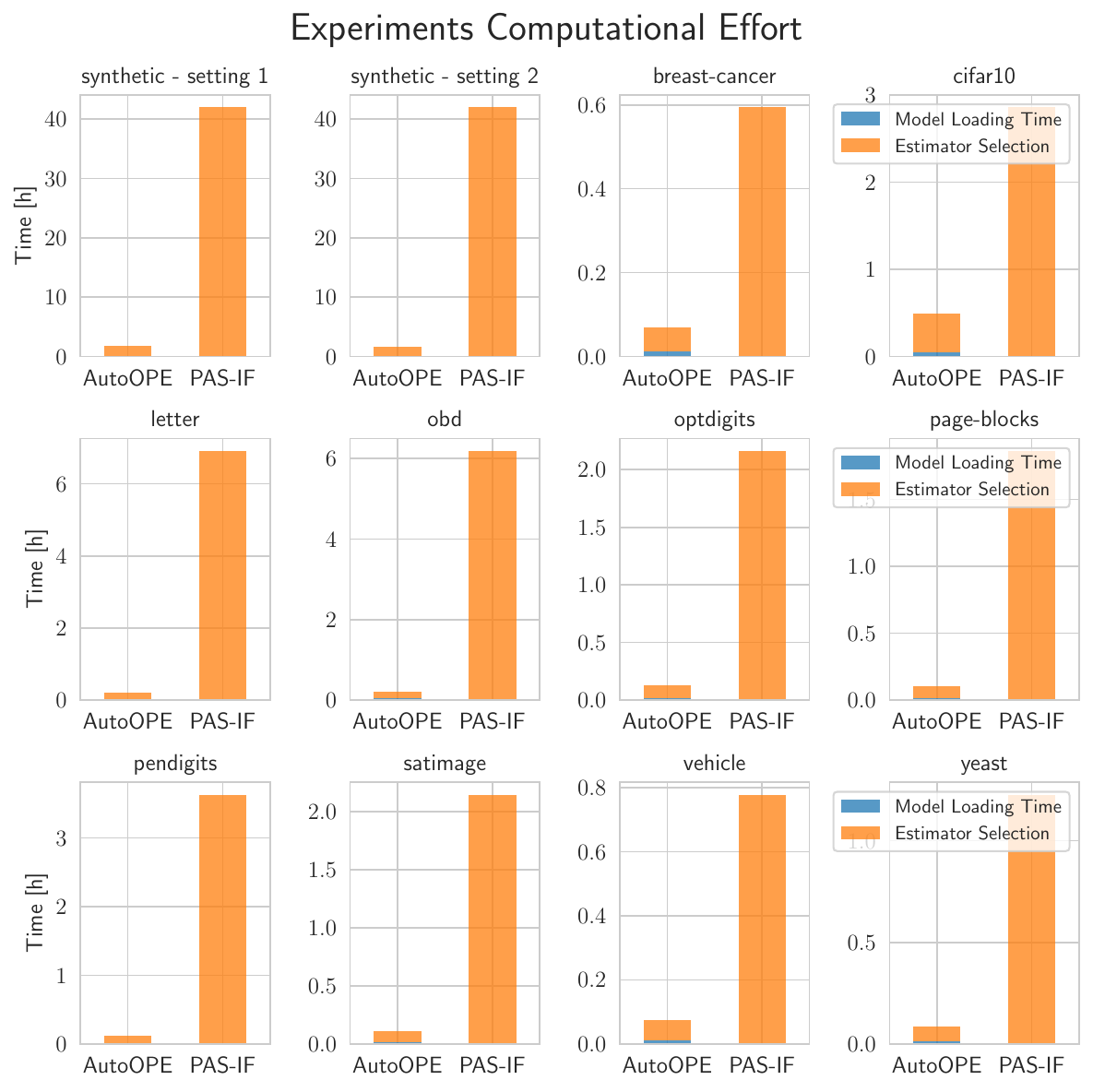}
    \caption{Comparison on the computational cost of \bbAutoOPE and \pasif for the Estimator Selection task.
    }
    \label{fig:comp-time-exp-all}
\end{minipage}
\end{figure}

Instead, the experiments described in Section \ref{sec:exp} require some hardware accelerator to speed-up the training of the neural network used by \pasif. Since the training is very demanding, we adopted the powerful \texttt{g6.12xlarge} instance, that owns four GPUs with 24GB of dedicated memory each, 48 CPUs and 192GB of RAM. Additionally, the \pasif code has undergone modifications to enable parallel computation wherever feasible, thereby reducing computational overhead during experimentation. In these experiments, it is apparent the high computational cost of \pasif, as shown in Figure \ref{fig:comp-time-exp-all}.

\begin{figure*}
    \centering
    \subfloat{\includegraphics[scale=0.6]{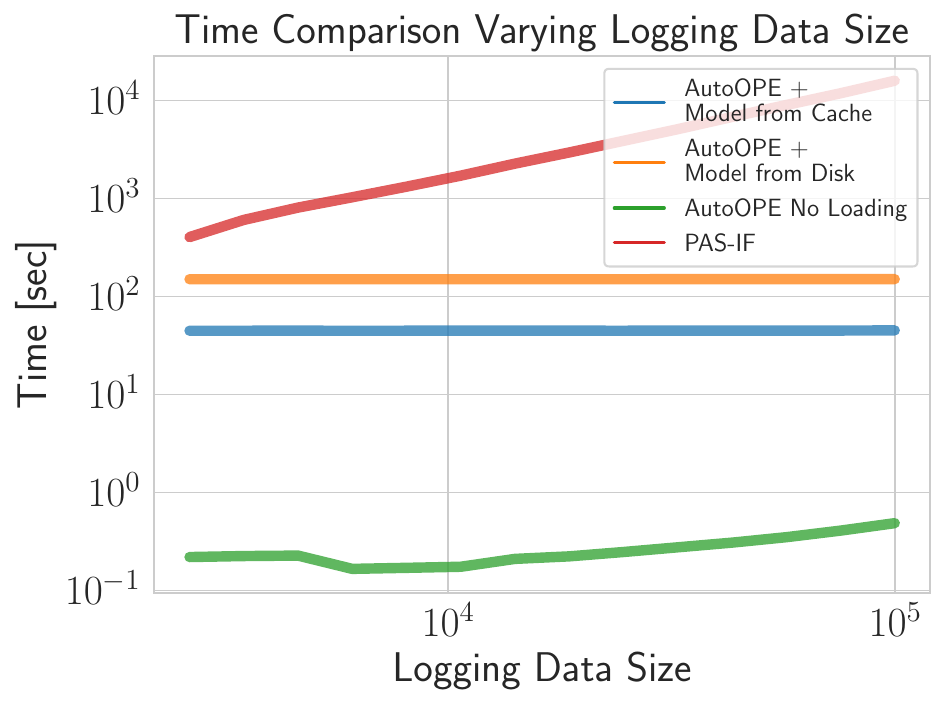}}
    \caption{Computational effort of \pasif and \bbAutoOPE in one single synthetic OPE task on \texttt{g6.12xlarge} AWS instance, with varying logging dataset dimension. The \bbAutoOPE is tested on two conditions, with the model loaded from cache, or from disk.}
    \label{fig:scaling-synt-exp}
\end{figure*}

We conduct an additional experiment to test the two Estimator Selection methods on one single synthetic OPE task, so that the maximum level of parallelization could be reached during the execution of \pasif.
Another factor that affects the time performance is the logging data size, so we perform this experiment different times, varying the logging dataset size. 
Concerning \bbAutoOPE instead, the model loading time is not negligible, so we test the method under two conditions: in the first condition the black-box model is already present in cache, while in the second the model need to be loaded from disk. The results are reported in Figure \ref{fig:scaling-synt-exp} and show that even considering the parallelized \pasif version and the model loading time for \bbAutoOPE, the time required by \pasif to perform the Estimator Selection task exceeds the time needed for \bbAutoOPE to load the model from disk and to perform the same Estimator Selection task. This difference is even higher for larger data sizes.

\section{GROUND-TRUTH EVALUATION AND POLICY VALUE ESTIMATION}
\label{app:mse-gt}
\subsection{Open Bandit Dataset}
With Open Bandit Dataset, the estimated ground-truth MSE of an estimator $\hat{V}_m$ predicting the evaluation policy value of $\pi_e$, can be computed using as ground-truth policy value estimate of $\pi_e$, the average of the counterfactual rewards of the evaluation dataset $\mathcal{D}_{gt}$ collected under the evaluation policy $\pi_e$. In the experiment outlined in Section \ref{subsubsec:obd}, $\pi_e$ is the Bernoulli Thompson Sampling policy. To ensure accurate ground-truth policy-value estimation, $\mathcal{D}_{gt} \coloneqq \left\{x_t, a_t, r_t, \left\{ \pi_e(a|x_t) \right\}_{a \in \mathcal{A}} \right\}_t^{n_{gt}}$ needs to contain a substantial number of rounds, as in this case, in which $n_{gt} = 12,357,200$.
The following expression describes the estimated ground-truth MSE of an estimator:

\begin{equation}
    \text{MSE}(\hat{V}_m(\pi_e; \task)) = \left(\hat{V}_m(\pi_e; \task) - \frac{1}{n_{gt}} \cdot \sum_{r_t \in \mathcal{D}_{gt}} r_t\right)^2
\end{equation}

\subsection{UCI Datasets and CIFAR-10}
In the experiment reported in Appendix \ref{app:cifar10} and in the experiments in Section \ref{subsubsec:uci}, we were able to compute the policy value of $\pi_e$ in the same way, since we use classification datasets. The policy value $\pie$ is computed as a true counterfactual expected reward on the logging dataset: in fact, the reward associated with each action and context in the logging data is deterministic (the reward is 1 for the correct class label/action relative to the context, and 0 for all other labels/actions). In this experiment the evaluation dataset can be defined as $\mathcal{D}_{gt} \coloneqq \left\{x_t, \left\{r_{t,v}\right\}_v^{|\mathcal{A}|}, \left\{\pi_e(a|x_t)\right\}_{a \in \mathcal{A}} \right\}_t^{n_b}$. The estimated ground-truth MSE of an estimator $\hat{V}_m$ predicting the evaluation policy value of $\pi_e$ with CIFAR-10 and UCI data (transformed in Contextual Bandit data as described in Appendix \ref{app:cifar10} and in Section \ref{subsubsec:uci}), is computed as described by the following expression:

\begin{equation}
    \text{MSE}(\hat{V}_m(\pi_e; \task)) = \left(\hat{V}_m(\pi_e; \task) - \frac{1}{n_b} \cdot \sum_t^{n_b} \sum_v^{|\mathcal{A}|} \pi_e(a_v|x_t) \cdot r_{t,v}\right)^2
\end{equation}

\subsection{Synthetic Data Experiments}
The procedure to estimate the ground-truth MSE of an estimator $\hat{V}_m$ predicting the evaluation policy value of $\pi_e$ with synthetic data, leverages on the possibility to produce synthetically a really big number of rounds under $\pi_e$. The evaluation data under policy $\pi_e$ are generated obtaining a dataset of this type: $\mathcal{D}_{gt} \coloneqq \left\{x_t, a_t, r_t, \left\{\pi_e(a|x_t)\right\}_{a \in \mathcal{A}}\right\}_t^{n_{gt}}$, and this process is repeated different times.
Furthermore, owing to the availability of expected rewards for each context-action pair in synthetic data, the policy value is computed as the weighted average of expected rewards, weighted by the action distribution. The whole procedure is described by Algorithm \ref{alg:gt-synt}.

\begin{algorithm}
\small
\caption{Ground-Truth MSE Estimation of OPE Estimators in Synthetic Data}
\begin{algorithmic}[1]
\REQUIRE number of data generations $n_{e}$, $\beta_e$ and a synthetic policy evaluation function $f_{\pi_e}$ to define $\pi_e$, $\beta_b$ and a synthetic policy evaluation function $f_{\pi_b}$ to define $\pi_b$, $n_b$ rounds to sample under $\pi_b$, a big number of rounds $n_{gt}$ to sample under $\pi_e$, synthetic reward function $q$, OPE estimator $\hat{V}_m$, space of available action $\mathcal{A}$
\ENSURE Estimated ground-truth policy value of $\pi_e,\ \hat{V}_{gt}(\pi_e)$
\FOR{$s = 1,\ldots,n_{e}$}
    \STATE $\mathcal{D}^{(s)}_{gt} \coloneqq \left\{x_t, a_t, r_t, \left\{f_{\pi_e}(x_t, a; s), q(x_t, a; s)\right\}_{a\in\mathcal{A}}\right\}_t^{n_{gt}} \leftarrow \mathrm{Generate} (\beta_e, f_{\pi_e}, q, n_{gt}, s)$ 
    \STATE \begin{varwidth}[t]{\linewidth}
    $\mathcal{T}^{(s)}_{ope} \coloneqq \left\{x_t, a_t, r_t, \left\{f_{\pi_b}(x_t, a; s), f_{\pi_e}(x_t, a; s), q(x_t, a; s)\right\}_{a\in\mathcal{A}}\right\}_t^{n_{b}} \leftarrow$ \par
    \hskip\algorithmicindent $\mathrm{Generate} (\beta_b, \beta_e, f_{\pi_b}, f_{\pi_e}, q, n_b, s)$
    \end{varwidth}
\ENDFOR 
\STATE $\text{MSE}(\hat{V}_m(\pi_e)) \leftarrow \frac{1}{n_{e}} \cdot \sum_s^{n_{e}} (\hat{V}_m(\pi_e; \task^{(s)}) - \frac{1}{n_{gt}} \cdot \sum_t^{n_{gt}} \sum_{a_j \in \mathcal{A}} f_{\pi_e}(x_t, a_j; s) \cdot q(x_t, a_j; s))^2$
\end{algorithmic}
\label{alg:gt-synt}
\end{algorithm}

In Appendix \ref{app:synt-exp} are specified mostly of the input parameters for \textbf{Logging 1} and \textbf{2} used in Algorithm \ref{alg:gt-synt}, a part from $n_{e},\ n_{gt}$, set with values $n_{e} = 100,\ n_{gt} = 1,000,000$.

With an improved performance in Estimator Selection, we improve the quality of OPE. This may be beneficial to society, allowing decision-makers to test policies that could be potentially dangerous without deploying them on a real system with improved reliability. Furthermore, using a pre-trained and not computationally expensive model can be beneficial in terms of environmental impacts.

\end{document}